\documentclass[times,authoryear]{elsarticle}

\usepackage{lineno,hyperref}
\modulolinenumbers[5]

\usepackage{framed,multirow}

\usepackage{amssymb}
\usepackage{latexsym}

\usepackage{url}
\usepackage{xcolor}
\usepackage{amsmath}
\usepackage{url}            
\usepackage{booktabs}       
\usepackage{amsfonts}       
\usepackage{nicefrac}       
\usepackage{microtype}      
\usepackage{graphicx}
\usepackage{caption}
\usepackage{subcaption}
\usepackage{lipsum}
\usepackage{xcolor}
\usepackage[noadjust]{cite}
\definecolor{newcolor}{rgb}{.8,.349,.1}

\def \ie {\emph{i.e.}}

\newcommand{\rev}[1]{\textcolor{black}{#1}}

\journal{Neural Networks}

\begin{document}

%
%

\begin{frontmatter}

\title{Working Memory Connections for LSTM}

\author[1]{Federico Landi\corref{cor1}} 
\cortext[cor1]{Corresponding author: 
  }
\ead{federico.landi@unimore.it}
\author[1]{Lorenzo Baraldi}
\author[1]{Marcella Cornia}
\author[1]{Rita Cucchiara}

\address[1]{Department of Engineering ``Enzo Ferrari'', University of Modena and Reggio Emilia, Modena, Italy}


\begin{abstract}
Recurrent Neural Networks with Long Short-Term Memory (LSTM) make use of gating mechanisms to mitigate exploding and vanishing gradients when learning long-term dependencies. For this reason, LSTMs and other gated RNNs are widely adopted, being the standard \textit{de facto} for many sequence modeling tasks. Although the memory cell inside the LSTM contains essential information, it is not allowed to influence the gating mechanism directly. In this work, we improve the gate potential by including information coming from the internal cell state. The proposed modification, named Working Memory Connection, consists in adding a learnable nonlinear projection of the cell content into the network gates. This modification can fit into the classical LSTM gates without any assumption on the underlying task, being particularly effective when dealing with longer sequences.
Previous research effort in this direction, which goes back to the early 2000s, could not bring a consistent improvement over vanilla LSTM.
As part of this paper, we identify a key issue tied to previous connections that heavily limits their effectiveness, hence preventing a successful integration of the knowledge coming from the internal cell state. We show through extensive experimental evaluation that Working Memory Connections constantly improve the performance of LSTMs on a variety of tasks. Numerical results suggest that the cell state contains useful information that is worth including in the gate structure.
\end{abstract}

\begin{keyword}
Long Short-Term Memory Networks\sep Cell-to-Gate Connections\sep Gated RNNs \sep Language Modeling \sep Image Captioning
\end{keyword}

\end{frontmatter}


\section{Introduction}
\label{sec:introduction}
Recurrent Neural Networks (RNNs)~\citep{Elman1990FindingSI,Rumelhart1986LearningRB} are a family of architectures that process sequential data by means of internal hidden states. The set of parameters of the network is shared across time steps, allowing the RNN to process inputs of variable length. 
As RNNs suffer from the so-called exploding and vanishing gradient problem (EVGP)~\citep{bengio1993problem,hochreiter1991untersuchungen}, which hinders the learning of long-term dependencies~\citep{bengio1994learning, pascanu2013difficulty}, previous works have proposed to enrich the recurrent cell with gating mechanisms~\citep{hochreiter1997long,jing2019gated}.
For instance, Long Short-Term Memory networks (LSTMs)~\citep{hochreiter1997long} use gates to control the information flow towards and from the memory cell and to regulate the forgetting process~\citep{gers2000forget}.
LSTMs are adopted in a wide number of tasks, such as neural machine translation~\citep{bahdanau2015neural,sutskever2014sequence}, speech recognition~\citep{graves2013speech}, and also vision-and-language applications like image and video captioning~\citep{vinyals2015show,xu2015show,baraldi2017hierarchical}. 

In this paper, we propose a novel cell-to-gate connection that modifies the classic LSTM block. Our formulation is general and improves LSTM overall performance and training stability without any particular assumption on the underlying task.
In the \textit{vanilla} LSTM formulation, the gates are controlled by the current input of the block and its previous output, which acts as the hidden state for the network. The long-term memory cell, instead, is employed to store information during the forward pass and provides a safe path for back-propagating the error signal. 
We argue that the content stored in the memory cell could be useful to regulate the gating mechanisms, too.
\rev{The key element of our design is a connection between the memory cell and the gates with a protection mechanism that prevents the cell state from being exposed directly. We draw inspiration from the gated \textit{read} operation employed to reveal the cell content at the block output, and enrich it with a learnable projection.} In this way, the LSTM block can use the knowledge in the cell (acting as a long-term memory) to control the evolution of the whole network in the short-term.

A similar concept in cognitive psychology and neuroscience is the so-called \textit{working memory}~\citep{ericsson1995long}, a type of memory employed, for instance, \textit{to retain the partial results while solving an arithmetic problem without paper, or to combine the premises in a lengthy rhetorical argument}~\citep{hernandez2018neuroethics}. Although definitions are not unanimous, working memory is said to be a cognitive system acting as a third type of memory between long-term and short-term memory. Our connections share this characteristic with working memory. For this reason, we call them \textit{Working Memory Connections} (WMCs).

A first attempt to fuse the information of the cell in the gates was made with the design of \textit{peepholes}~\citep{gers2000recurrent}: direct multiplicative connections between the memory cell and the gates. This approach has not been largely adopted in literature, as recent studies report mixed results~\citep{Greff2017LSTMAS} and discourage their use. Since our idea recalls the rationale of peephole connections, we provide a large comparison with this previous work. By doing so, we point out the major issues in the peephole formulation that hinder effective learning and attest that WMCs do not suffer from the same problems.
In our experiments, we show that an LSTM equipped with Working Memory Connections achieves better results than comparable architectures, thus reflecting the theoretical advantages of their design. In particular, WMCs surpass vanilla LSTM and peephole LSTM in terms of final performances, stability during training, and convergence time. All these aspects testify the advantage in letting the cell state participate in the gating dynamics. In order to support our conclusions, we conduct a thorough experimental analysis covering a wide area of current research topics.

\rev{To sum up, our contribution is mainly three-fold. First, we present a modification of LSTM in which traditional gates are enriched with Working Memory Connections, linking the memory cell with the gates through a protection mechanism. Then, we demonstrate that exposing the LSTM internal state directly and without a proper protection yields unstable training dynamics that compromise the final performance.} Finally, we show the effectiveness of the proposed solution in a variety of tasks, ranging from toy problems with very long-term dependencies (adding problem, copy task, and sequential MNIST) to language modeling and image captioning.

\section{Related Work}
\label{sec:related}
Long Short-Term Memory networks~\citep{hochreiter1997long} aim to mitigate the exploding and vanishing gradient problem~\citep{hochreiter1991untersuchungen, bengio1994learning} with the use of gating mechanisms. Since its introduction, LSTM has gained a lot of attention for its flexibility and efficacy in many different tasks.
To simplify the LSTM structure,~\citet{liu2020simplified} propose to exploit the content of the long-term memory cell in a recurrent block with only two gates. However, this model neglects the importance of the LSTM output. While this might be useful for simple tasks, it is unlikely to generalize to more complex settings.~\citet{arpit2018h} propose to modify the path of the gradients in order to stabilize training with a stochastic algorithm specific to LSTM optimization. This direction of work is not in contrast with our goal, and could possibly be integrated with our proposal since our connection does not require a specific setup to be optimized.
Among the LSTM variants, the Gated Recurrent Unit (GRU)~\citep{cho2014LearningPR,cho2014properties} is the most popular and common architecture~\citep{chung2014empirical}, and features a coupling mechanism between input and forget gates~\citep{Greff2017LSTMAS}.
A recent line of research aims to tailor the LSTM structure for specific tasks. For instance,~\citet{baraldi2017hierarchical} propose a hierarchical model for video captioning, while other works incorporate convolutional models into the LSTM structure~\citep{XIAO2020173, LI201841}.
While these works propose a modification of the LSTM towards a specific goal, we propose a general and powerful idea that adapts to a large set of different tasks.

Recently, models based on self-attention, such as Transformer architecture~\citep{vaswani2017attention} and its variants, are achieving state-of-art performances on many different tasks, and also for sequence modeling. For instance, language representations based on BERT~\citep{devlin2018bert} can be finetuned with an additional output layer to obtain state-of-art results on many language-based tasks. However, RNNs require much fewer parameters and operations to run than Transformer-based architectures and are still widely adopted.
Moreover, LSTMs still have a large market in embedded systems and edge devices for their low computational and memory requirements.

\section{Proposed Method}
\label{sec:method}
In this section, we present a complete overview of Working Memory Connections. First, we recall the LSTM equations. Second, we explain the modifications introduced in our design. Finally, we motivate the choices behind WMCs \textit{w.r.t.}~other approaches. Specifically, we identify key problems in previous cell-to-gate connections that hinder the learning process, and we show that the proposed solution does not suffer from these weaknesses.

\subsection{LSTM}
The core idea behind Long Short-Term Memory networks is to create a constant error path between subsequent time steps.
Being $\mathbf{x}_t$ the input vector at time $t$ we can write the rollout equations for a vanilla LSTM as:
\begin{align}
    &\mathbf{g}_t = \text{tanh}
                    (\mathbf{W}_{gx}\mathbf{x}_{t} +
                    \mathbf{W}_{gh} \mathbf{h}_{t-1} +
                    \mathbf{b}_g) \\
    &\mathbf{i}_t = \sigma
                    (\mathbf{W}_{ix}\mathbf{x}_{t} +
                    \mathbf{W}_{ih} \mathbf{h}_{t-1} +
                    \mathbf{b}_i)
                    \label{eq:i} \\
    &\mathbf{f}_t = \sigma
                    (\mathbf{W}_{fx}\mathbf{x}_{t} +
                    \mathbf{W}_{fh} \mathbf{h}_{t-1} +
                    \mathbf{b}_f)
                    \label{eq:f} \\
    &\mathbf{c}_t = \mathbf{f}_t \odot \mathbf{c}_{t-1} +
                    \mathbf{i}_t \odot \mathbf{g}_t \\
    &\mathbf{o}_t = \sigma
                    (\mathbf{W}_{ox}\mathbf{x}_{t} +
                    \mathbf{W}_{oh} \mathbf{h}_{t-1} +
                    \mathbf{b}_o)
                    \label{eq:o} \\
    &\mathbf{h}_t =  \mathbf{o}_t \odot \text{tanh}(\mathbf{c}_t).
                    \label{eq:h}
\end{align}
Here, $\mathbf{g}$ is the block input, $\mathbf{i}$, $\mathbf{f}$, and $\mathbf{o}$ are respectively the input, forget, and output gates, $\mathbf{c}$ represents the memory cell value, and $\mathbf{h}$ is the block output. In this notation, $\sigma$ is the sigmoid function and $\odot$ denotes element-wise Hadamard product.
In its first formulation~\citep{hochreiter1997long}, LSTM did not include the multiplicative forget gate. However, being able to forget about past inputs~\citep{gers2000forget} allows LSTM to tackle longer sequences while not hindering the back-propagation of the error signal.


\begin{figure*}[!t]
    \centering
    \includegraphics[height=3.8cm]{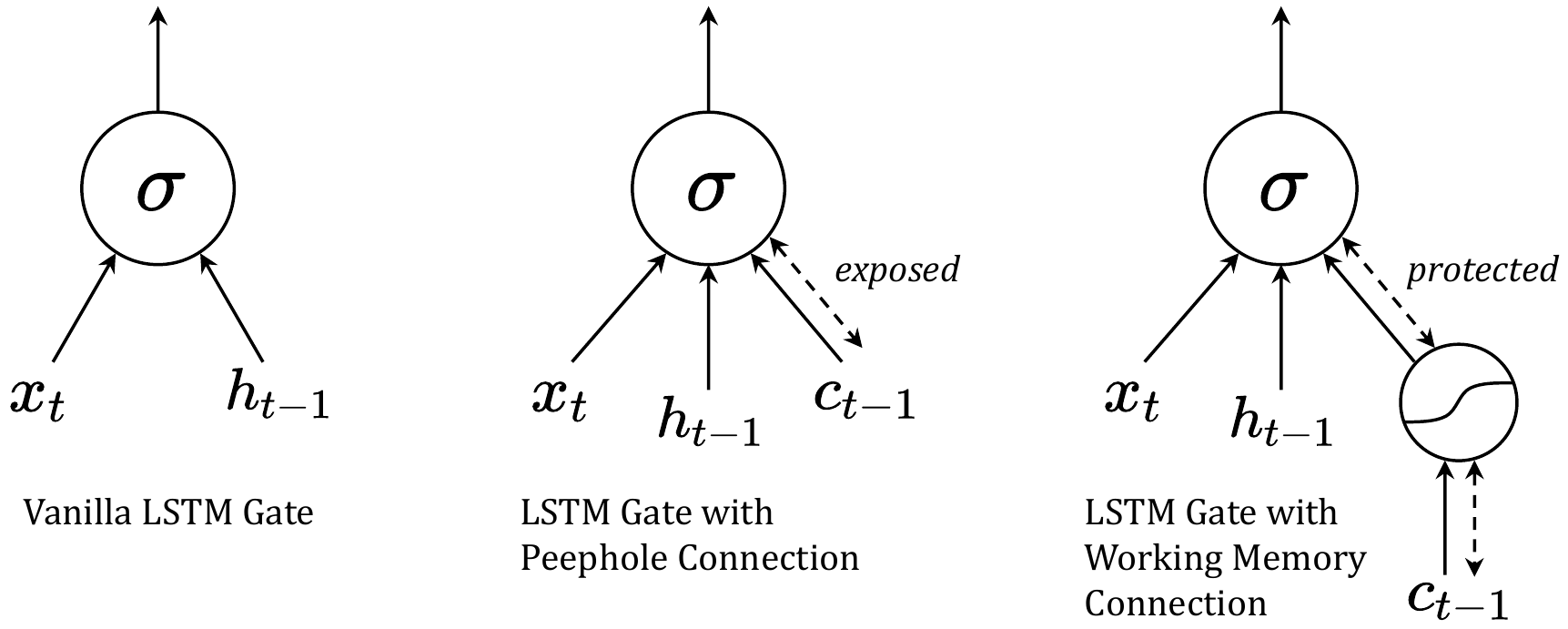}
    \caption{Comparison between a vanilla LSTM gate, a peephole connection, and a Working Memory Connection.}
\label{fig:1a}
\end{figure*}

\subsection{Working Memory Connections}
In the following, we introduce Working Memory Connections, which enable the memory cell to influence the value of the gates through a set of recurrent weights. Given a proper design for the connection, we argue that there is a practical advantage in letting the cell state influence the gating mechanisms in the LSTM block directly. In fact, the cell state $\mathbf{c}_t$ provides unique information about the previous time steps that are not present in $\mathbf{h}_t$. 
For instance, $\mathbf{h}_t$ may be close to zero as a consequence of the output gate saturating towards zero (see Eq.~\ref{eq:h}), while $\mathbf{c}_t$ may be growing and changing as a result of a sequence of input vectors. In that case, since the cell state cannot control the output gate, the LSTM block is forced to learn which particular value in the input vector is the marker that signals to open the output gate. Instead, with an appropriate connection strategy, the LSTM block could learn a mapping between the cell internal state and the gate values.

Our solution employs a set of recurrent weights  $W_{\star c}$, $\star \in \{\mathbf{i}, \mathbf{f}, \mathbf{o}\}$ and a nonlinear activation function to model a connection between memory cell and gates. The application of a non-linearity on the memory cell is coherent with the present LSTM structure: as it can be noticed from Eq.~\ref{eq:h}, a nonlinear activation function is applied to $\mathbf{c}_t$ before the Hadamard product with $\mathbf{o}_t$\footnote{Previous works~\citep{Greff2017LSTMAS} have also shown that removing this non-linearity leads to a significant loss in terms of performance.}. 
\rev{In light of the above-mentioned intuitions, we modify Eq.~\ref{eq:i},~\ref{eq:f}, and~\ref{eq:o} by exposing the cell state $\mathbf{c}_t$ at time $t$ through a protection mechanism as follows:} 
\begin{align}
    &\mathbf{i}_t = \sigma
                    (\mathbf{W}_{ix}\mathbf{x}_{t} +
                    \mathbf{W}_{ih} \mathbf{h}_{t-1} +
                    \text{tanh}(\mathbf{W}_{ic} \mathbf{c}_{t-1}) +
                    \mathbf{b}_i) 
                    \label{eq:pv2i} \\
    &\mathbf{f}_t = \sigma
                    (\mathbf{W}_{fx}\mathbf{x}_{t} +
                    \mathbf{W}_{fh} \mathbf{h}_{t-1} +
                    \text{tanh}(\mathbf{W}_{fc} \mathbf{c}_{t-1}) +
                    \mathbf{b}_f) 
                    \label{eq:pv2f} \\
    &\mathbf{o}_t = \sigma
                    (\mathbf{W}_{ox}\mathbf{x}_{t} +
                    \mathbf{W}_{oh} \mathbf{h}_{t-1} +
                    \text{tanh}(\mathbf{W}_{oc} \mathbf{c}_{t}) +
                    \mathbf{b}_o),
                    \label{eq:pv2o}
\end{align}
where $\mathbf{W}_{\star c} \mathbf{c}_t$ denotes a general linear transformation.

At a first glance, Working Memory Connections may seem redundant in the gate structure. In fact, $h_{t-1}$ depends from the value of $c_{t-1}$ (Eq.~\ref{eq:h}). This impression is misleading, as the proposed connections introduce two main aspects of novelty. First, the non-linear activation function operates on three different projections of the cell state, one for each gate type. Second, Eq.~\ref{eq:pv2o} shows that the connection on the output gate depends on $c_t$, rather than on $c_{t-1}$, hence allowing for a more responsive control of the output dynamics of the entire LSTM block.


\subsection{Advantages of Working Memory Connections}
\label{subsec:ph}
To formally motivate the improvement given by Working Memory Connections, we start by considering the local gradients of the gates in which the cell interaction is added. We limit our formal analysis to the input gate $\mathbf{i}_t$, but our reasoning can be generalized to $\mathbf{f}_t$ and $\mathbf{o}_t$.
If we denote by $\mathbf{\bar i}_t$ the argument of the sigmoid activation function (Eq.~\ref{eq:pv2i}) at time $t$:
\begin{align}
        \mathbf{\bar i}_t &= \mathbf{W}_{ix}\mathbf{x}_{t} +
                                     \mathbf{W}_{ih} \mathbf{h}_{t-1} +
                                     \text{tanh}(\mathbf{W}_{ic} \mathbf{c}_{t-1}) +
                                     \mathbf{b}_i ,
\end{align}
\rev{then the local gradient of the input gate $\mathbf{i}_t$ is expressed by:
\begin{align}
     \frac{\partial\mathbf{i}_t}{\partial\mathbf{\bar i}_t} &=
     \frac{\partial}{\partial\mathbf{\bar i}_t} \sigma \left(\, \mathbf{\bar i}_t \right) =
                          \textit{diag}\left[
                          \sigma\left(\, \mathbf{\bar i}_t \right) \odot
                          \left(\mathbf{1} - \sigma\left(\, \mathbf{\bar i}_t \right) \right)
                          \right],
\end{align}
where $\mathbf{1}$ denotes a vector of ones, and $\textit{diag}[\mathbf{x}]$ indicates a diagonal $\text{N} \times \text{N}$ matrix whose diagonal contains the N elements of vector $\mathbf{x}$.}

\rev{From here, we can easily derive the local gradients on the recurrent weights $\mathbf{W}_{ix}$, $\mathbf{W}_{ih}$, and $\mathbf{W}_{ic}$ at time $t$:
\begin{align}
    \frac{\partial\mathbf{i}_t}{\partial\mathbf{W}_{ix}} &= 
    \frac{\partial\mathbf{i}_t}{\partial\mathbf{\bar i}_t} \otimes \mathbf{x}_t , \\
    \frac{\partial\mathbf{i}_t}{\partial\mathbf{W}_{ih}} &= 
    \frac{\partial\mathbf{i}_t}{\partial\mathbf{\bar i}_t} \otimes \mathbf{h}_{t-1} , \\
    \frac{\partial\mathbf{i}_t}{\partial\mathbf{W}_{ic}} &=
    \delta \mathbf{\hat i}_t \otimes \mathbf{c}_{t-1} ,
    \label{eq:di1}
\end{align}
where $\otimes$ denotes the outer product of two vectors, and:
\begin{align}
    \delta \mathbf{\hat i}_t &= \frac{\partial\mathbf{i}_t}{\partial\mathbf{\bar i}_t} \odot
    \left( \mathbf{1} - \text{tanh}^2 (\,\mathbf{W}_{ic} \mathbf{c}_{t-1} ) \, \right).
    \label{eq:di2}
\end{align}}
Now, let's consider what happens as $t$ grows: we observe that $\mathbf{x}_t$ and $\mathbf{h}_t$ are bounded to a limited interval. In particular, $\mathbf{x}_t$ is a sample of the input data, and $\mathbf{h}_t$ is bounded in the interval $[-1, 1]$ by construction. Instead, the cell $\mathbf{c}_t$ can grow linearly with the number of recursive steps, making its domain extremely task-dependent.
This is a well-known problem, which motivated the introduction of the forget gate in the original LSTM structure~\citep{gers2000forget}. Despite this, the range of possible values of $c_t$ cannot be restricted to a fixed domain.
The hyperbolic tangent non-linearity helps to avoid an excessive influence of the unbounded cell state in the gate mechanics, hence preventing unwanted saturation. As it can be seen in Eq.~\ref{eq:pv2i},~\ref{eq:pv2f}, and~\ref{eq:pv2o}, the term related to the cell state is bounded in the interval $[-1, 1]$. Additionally, it helps screen the connection weights $\mathbf{W}_{\star c}$ from unstable updates.

Even if $\mathbf{c}_t$ grew linearly with the number of time steps, its influence on the sigmoid argument would be mitigated, and it could not take the sigmoid function into its saturated regime against the other two terms driven by $\mathbf{x}$ and $\mathbf{h}$ respectively.
On the other hand, the growth of the cell state would push the hyperbolic tangent towards its own saturated regime. This behavior helps protect the weight matrix employed in the connection from unstable updates.

\smallbreak
\noindent\textbf{Peephole Connections and their Limitations.}
We now turn our attention to a related connection, namely the peephole connection~\citep{gers2000recurrent}, which is no longer common in the LSTM formulation.
Peephole connections were introduced by Gers and Schmidhuber in~\citep{gers2000recurrent}, and enrich the LSTM equations with recurrent weights $\mathbf{W}_{\star c}$, $\star \in \{\mathbf{i}, \mathbf{f}, \mathbf{o}\}$:
\begin{align}
    &\mathbf{i}_t = \sigma
                    (\mathbf{W}_{ix}\mathbf{x}_{t} +
                    \mathbf{W}_{ih} \mathbf{h}_{t-1} +
                    \mathbf{W}_{ic} \mathbf{c}_{t-1} +
                    \mathbf{b}_i)
                    \label{eq:ph_i} \\
    &\mathbf{f}_t = \sigma
                    (\mathbf{W}_{fx}\mathbf{x}_{t} +
                    \mathbf{W}_{fh} \mathbf{h}_{t-1} +
                    \mathbf{W}_{fc} \mathbf{c}_{t-1} +
                    \mathbf{b}_f) \\
    &\mathbf{o}_t = \sigma
                    (\mathbf{W}_{ox}\mathbf{x}_{t} +
                    \mathbf{W}_{oh} \mathbf{h}_{t-1} +
                    \mathbf{W}_{oc} \mathbf{c}_{t} +
                    \mathbf{b}_o),
\end{align}
with $\mathbf{W}_{\star c}$ generally constrained to be diagonal~\citep{graves2013generating, Greff2017LSTMAS}. While this formulation allows for a more precise control of the gates, there are two issues that limit its effectiveness.
\rev{In this case, the local gradient at time $t$ is expressed by:
\begin{align}
    \frac{\partial\mathbf{i}_t}{\partial\mathbf{\bar i}_t} &=
    \frac{\partial}{\partial\mathbf{\bar i}_t} \sigma \left(\, \mathbf{\bar i}_t \right) =
                          \textit{diag}\left[
                          \sigma\left(\, \mathbf{\bar i}_t \right) \odot
                          \left( \mathbf{1} - \sigma\left(\, \mathbf{\bar i}_t \right) \right)
                          \right] ,
\end{align}
}
with $\mathbf{\bar i}_t$ being the argument of the sigmoid function in Eq~\ref{eq:ph_i}:
\begin{align}
    \mathbf{\bar i}_t &= \mathbf{W}_{ix}\mathbf{x}_{t} +
                                     \mathbf{W}_{ih} \mathbf{h}_{t-1} +
                                     \mathbf{W}_{ic} \mathbf{c}_{t-1} +
                                     \mathbf{b}_i .
\end{align}
\rev{In light of this difference, Eq.~\ref{eq:di1} and~\ref{eq:di2} become:
\begin{align}
    \frac{\partial\mathbf{i}_t}{\partial\mathbf{W}_{ic}} &= 
    \frac{\partial\mathbf{i}_t}{\partial\mathbf{\bar i}_t} \otimes \mathbf{c}_{t-1}  .
    \label{eq:deltaw}
\end{align}
}
\begin{figure*}[!t]
    \centering
    \includegraphics[height=4.2cm]{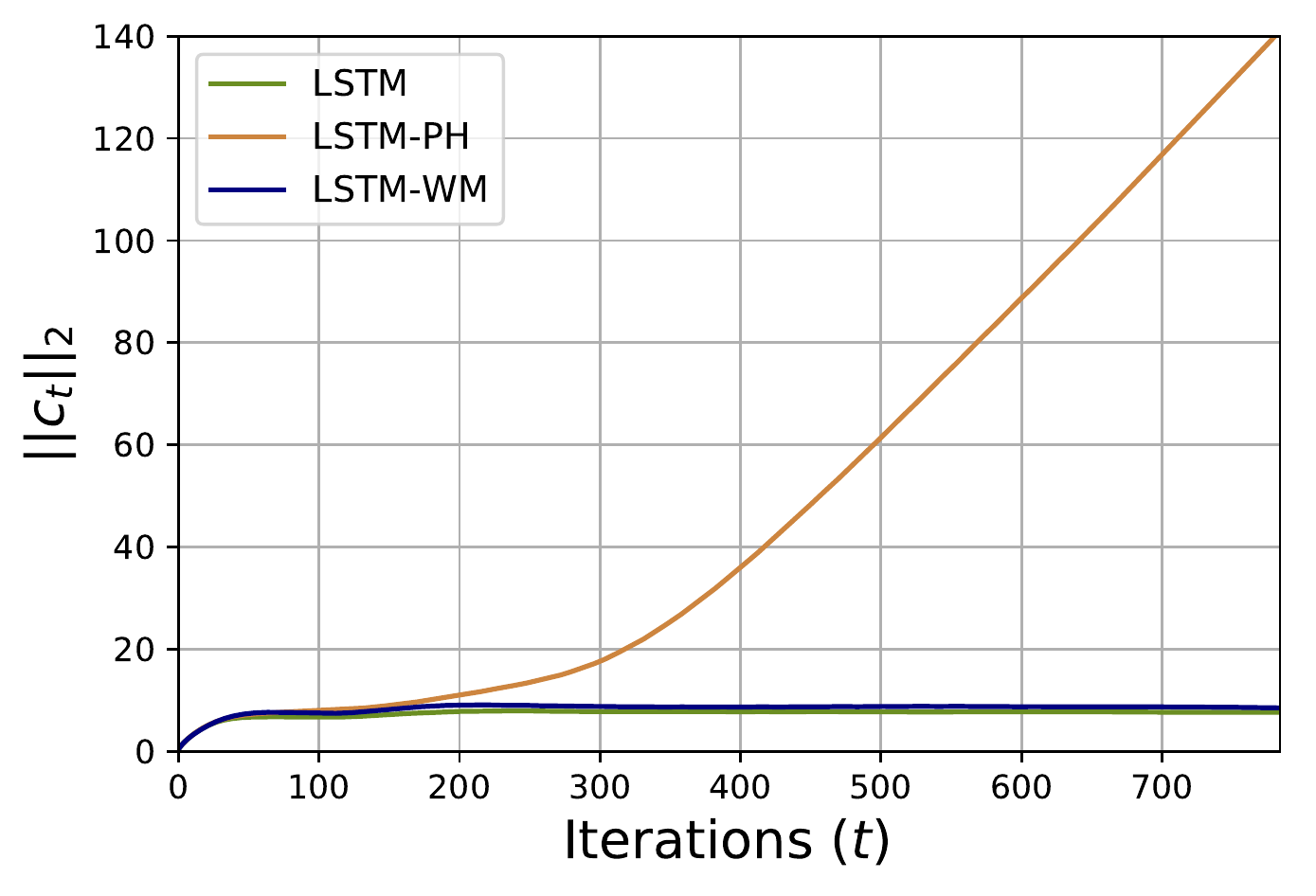}
    \caption{The cell state $c_t$ may grow linearly with the number of time steps. Peephole connections directly expose $c_t$, creating a key issue (b). Data for this plot is taken from the first training iterations of the sequential MNIST (see \textsection~\ref{sec:smnist}).}
\label{fig:1b}
\end{figure*}

\rev{We observe that, both in Eq.~\ref{eq:pv2i} and in Eq~\ref{eq:ph_i}, the magnitude of the product $\mathbf{W}_{ic} \mathbf{c}_{t-1}$ can in principle grow unbounded.
The activation function introduced in WMCs squashes this term into a closed bounded interval. In peephole connections, hovewer, this term is added inside the gate without an adequate protection (see Fig.~\ref{fig:1a}).
The result is that, in the peephole formulation, the sigmoid function applied immediately after could be pushed towards its saturating regime independently from the value of $\mathbf{x}_t$ and $\mathbf{h}_t$.
In theory, the LSTM block can recover from this situation by setting all the weights in the peephole connection to $0$, but in practice this might not happen if the sigmoid gate is saturated most of the time. Even if the two other summands can compensate for the growth of $\mathbf{W}_{\star c}\mathbf{c}$, hence letting gradients flow through the gate, there is still a key issue that hinders learning. In fact, as shown in Eq. \ref{eq:deltaw}, the gradients on the recurrent peephole weights grow linearly with $\mathbf{c}$, making updates unstable.}

To exemplify this behavior, we report the Euclidean norm of $\mathbf{c}_t$ during the early training stages in Fig.~\ref{fig:1b}. After a small number of time steps, the content of the cell floods the gates of the peephole LSTM. A possible consequence would be that both the input and the forget gates would saturate towards $1$. In our example, this aspect leads to an additional and uncontrolled growth of the magnitude of $\mathbf{c}_t$. 
As it can be seen, Working Memory Connections exhibit a much more regular behavior than peepholes and can prevent the uncontrolled growth of the memory cell.

\section{Experiments and Results}
\label{sec:results}
\begin{figure*}[t!]
\setlength{\tabcolsep}{.3em}
\centering
\renewcommand{\arraystretch}{0.5}
\begin{tabular}{cc}
    \scriptsize\textbf{Adding Problem -- T=200} &  \scriptsize\textbf{Adding Problem -- T=400}\\
    \includegraphics[height=0.315\linewidth]{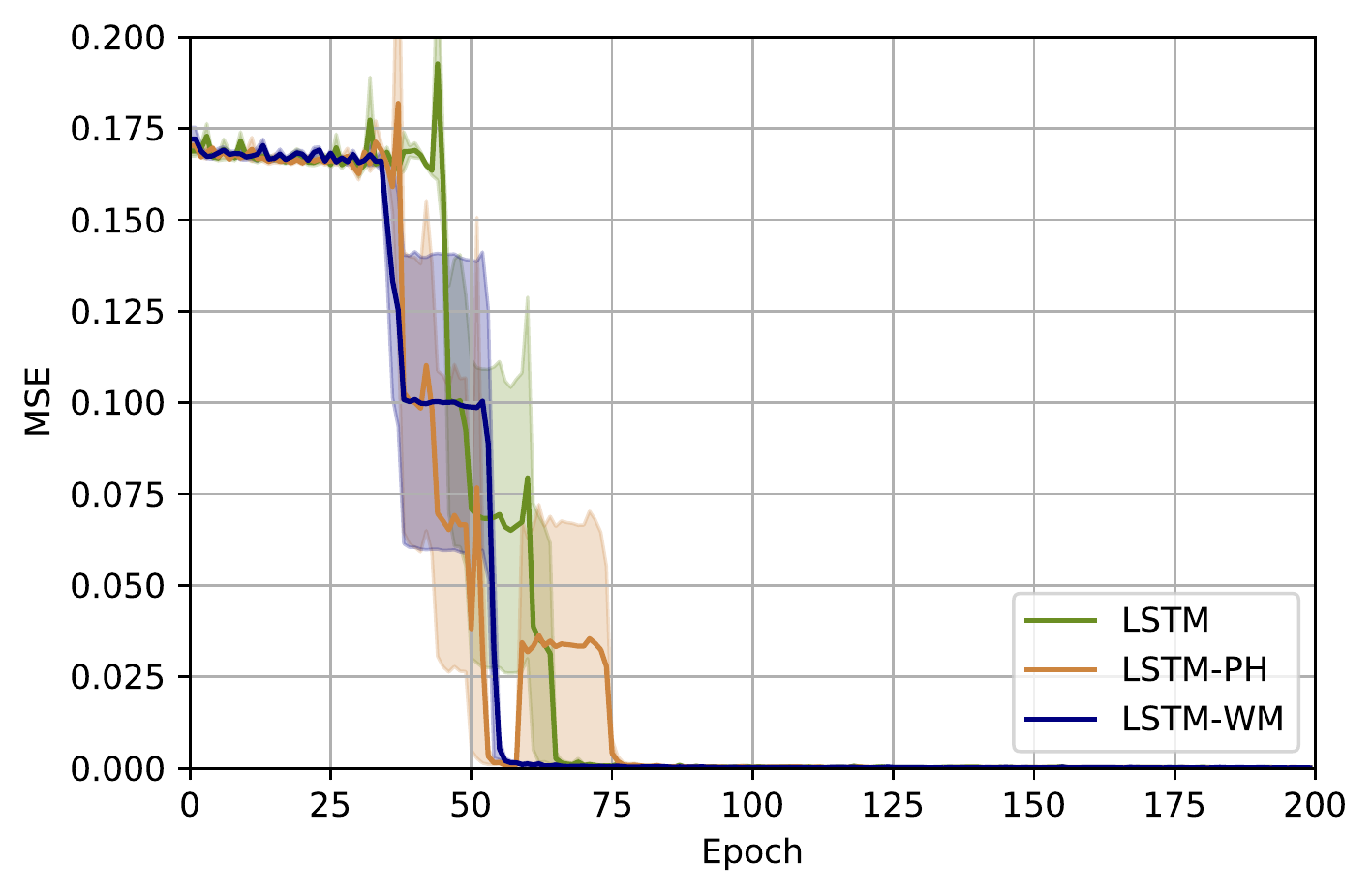}
     &  
     \includegraphics[height=0.315\linewidth]{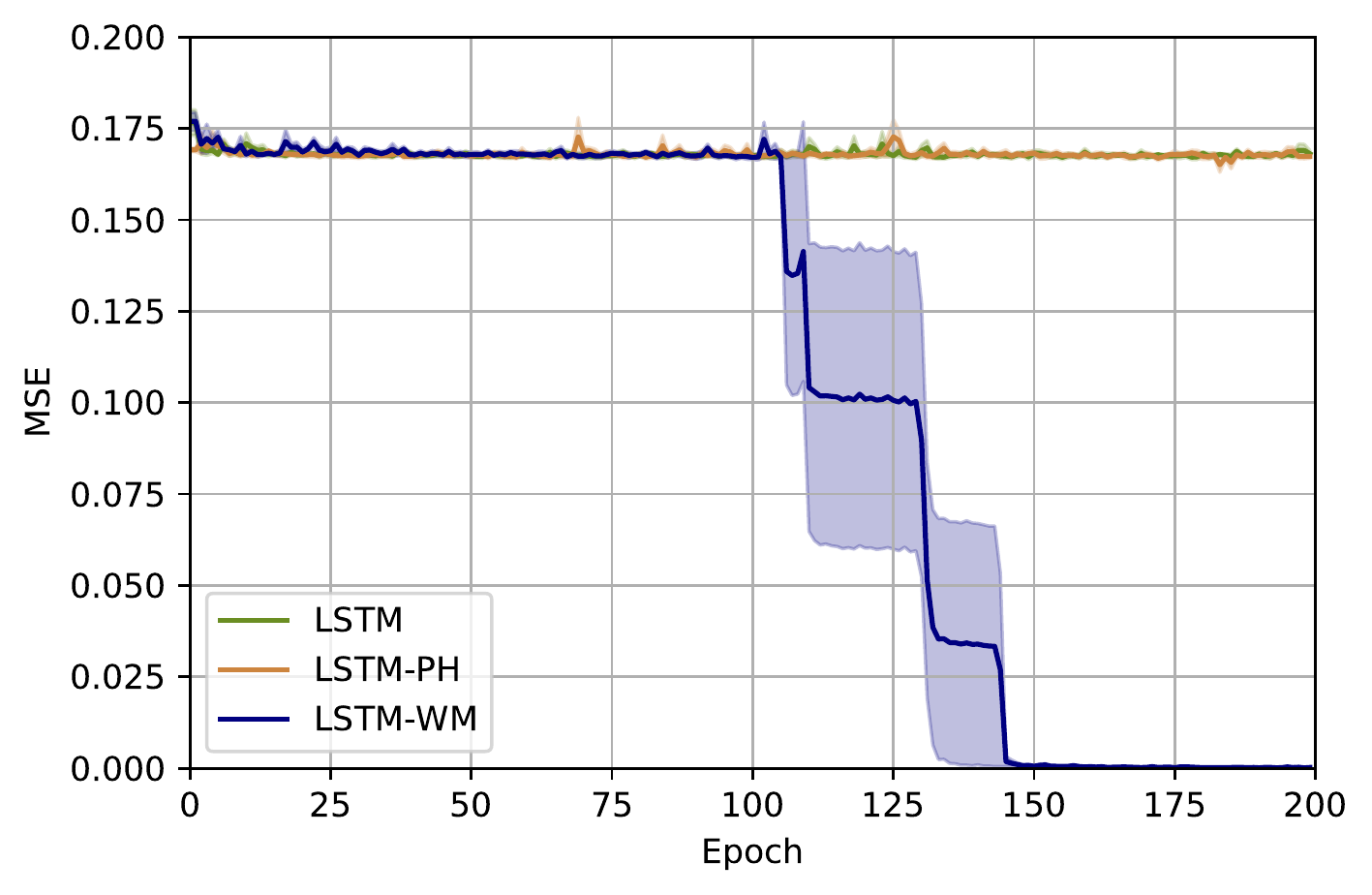}
     \\
    \scriptsize\textbf{Copying Task -- T=100} & \scriptsize\textbf{Copying Task -- T=200}\\
    \includegraphics[height=0.315\linewidth]{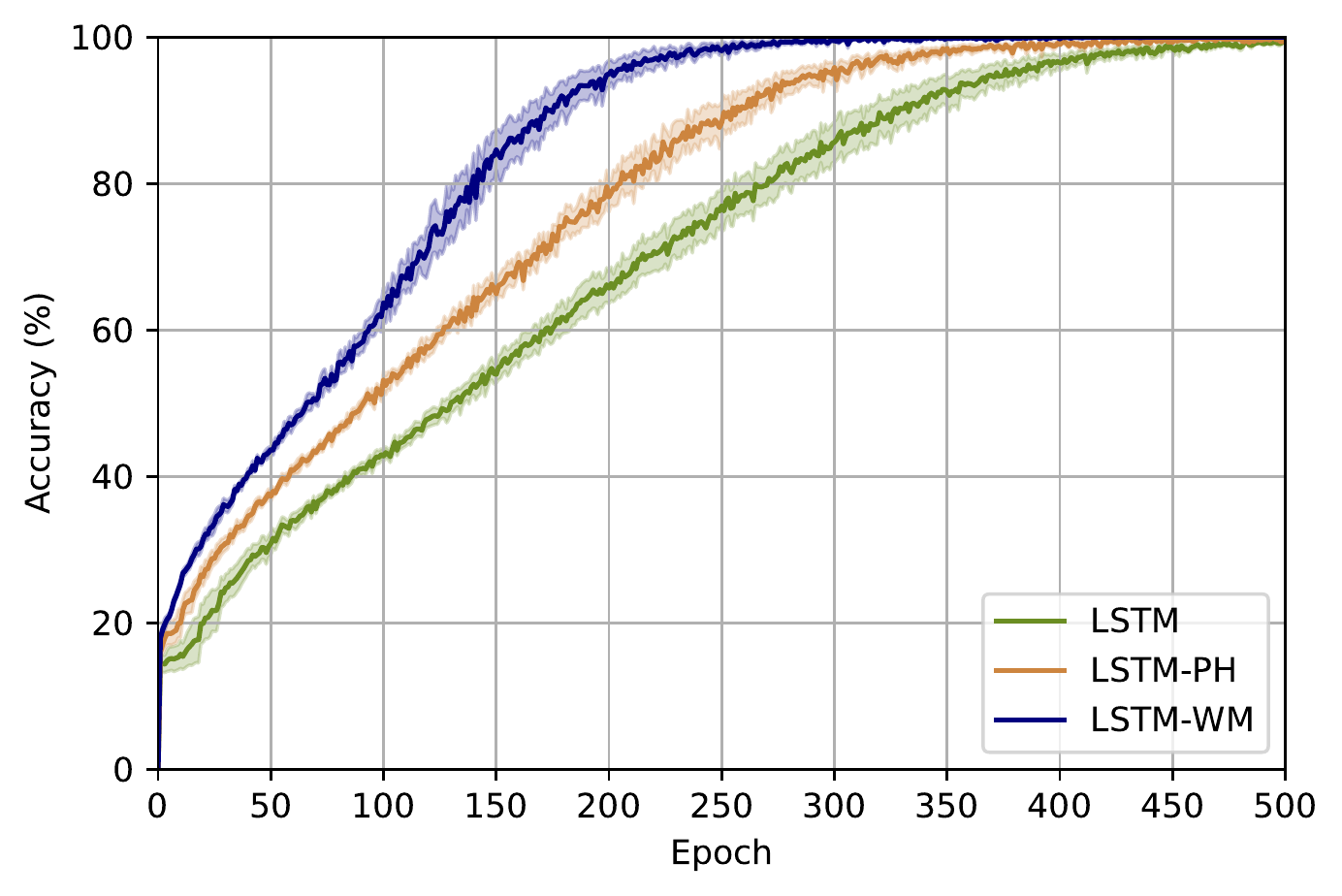} 
    &
   \includegraphics[height=0.315\linewidth]{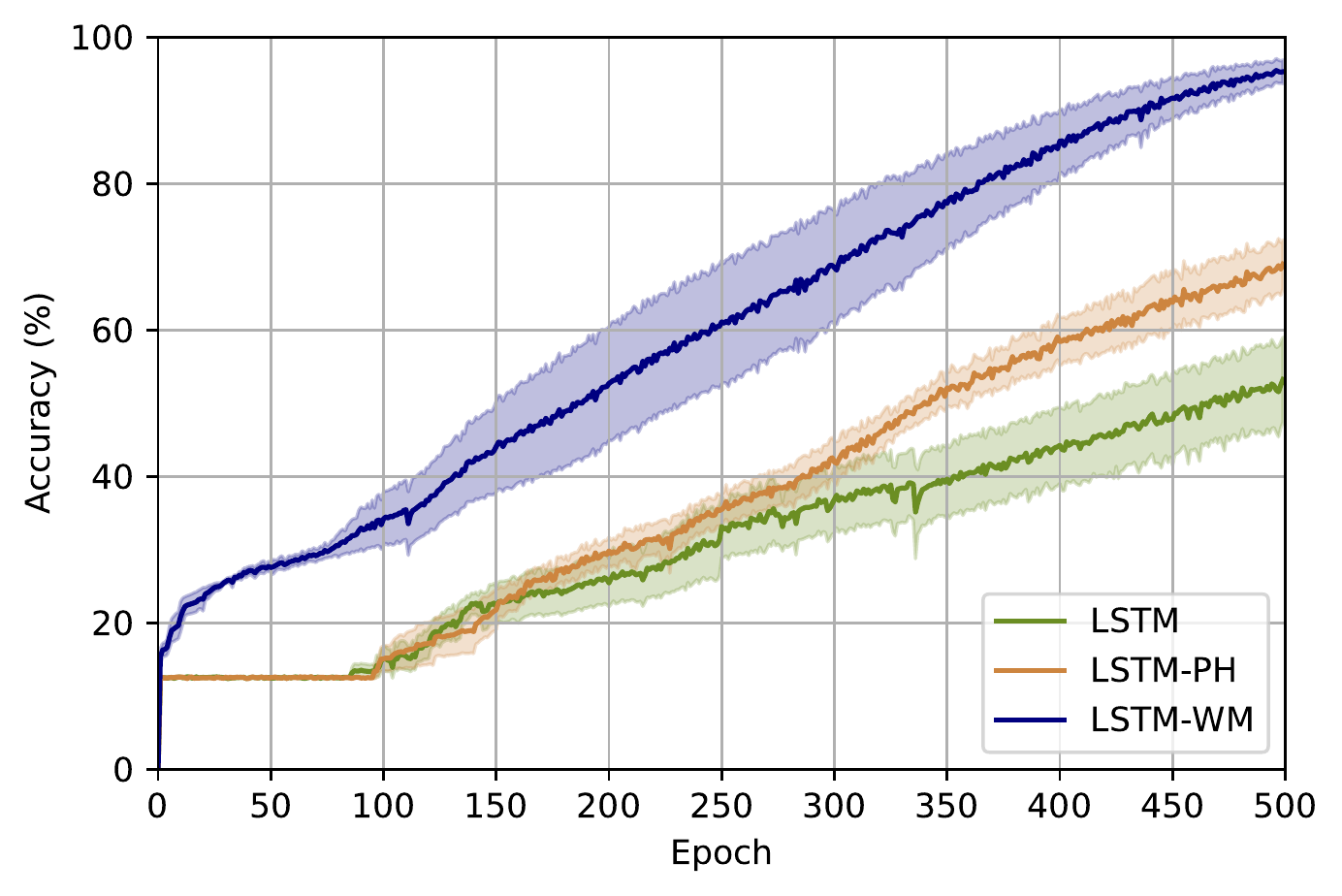}
    \\
    \scriptsize\textbf{Sequential MNIST} & \scriptsize\textbf{Permuted MNIST} \\
    \includegraphics[height=0.315\linewidth]{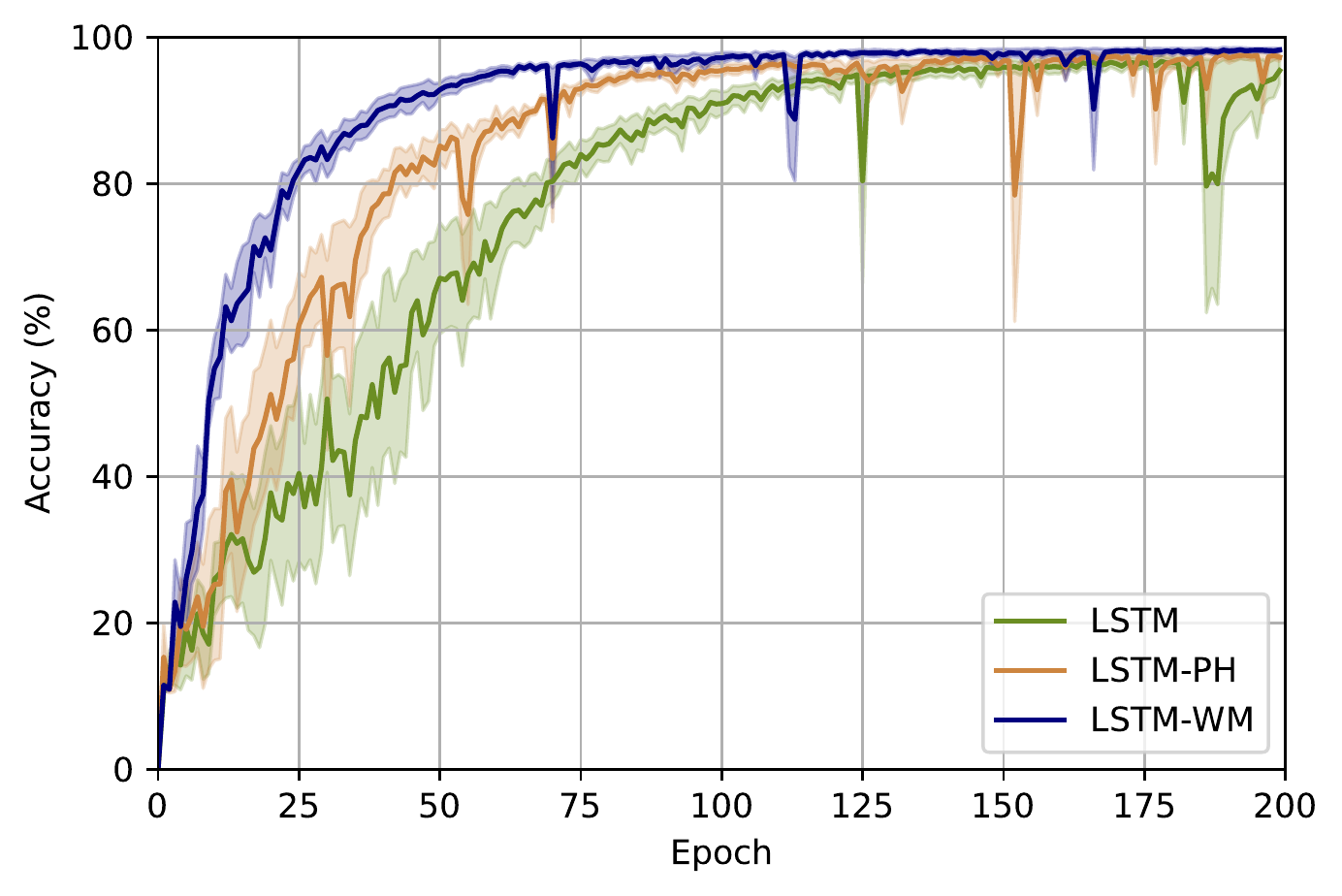} 
     &  
    \includegraphics[height=0.315\linewidth]{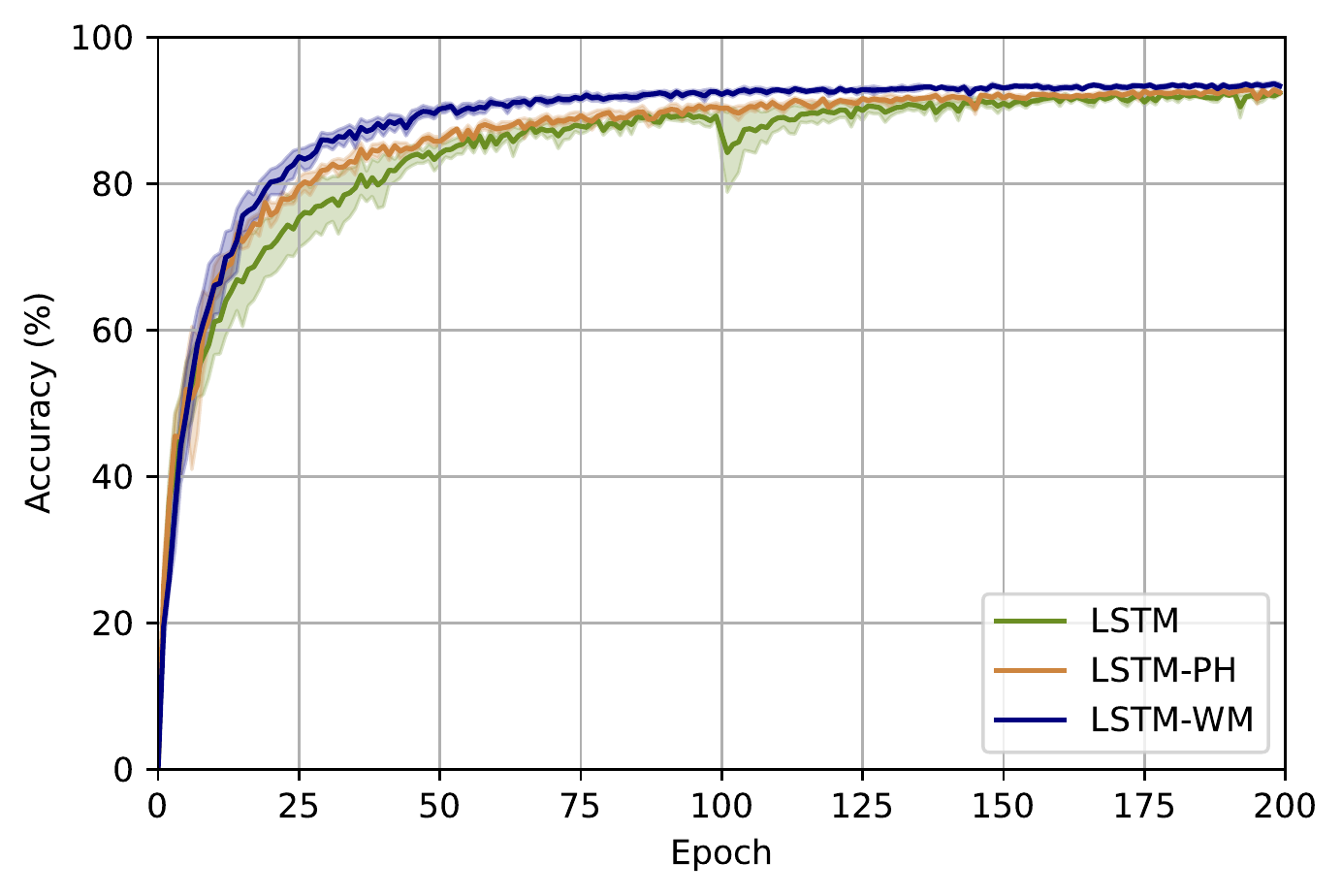}
\end{tabular}
\caption{Comparison among traditional LSTM, the proposed LSTM with working memory connections, and peephole LSTM. We investigate three different tasks: the adding problem (top), the copying task (center), and the sequential/permuted MNIST (bottom). In all the plots, shading indicates the standard error of the mean.}
\label{fig:plots}
\end{figure*}

The effectiveness of Working Memory Connections and their general benefits can be appreciated in many different tasks. The proposed experiments cover a wide area of applications: two different toy problems, digit recognition, language modeling, and image captioning. While the analysis on simple tasks helps to clarify the inherent advantages of the proposed approach, results on more challenging real-world applications motivate a wider adoption of our novel connections, especially for long sequences.
We compare our model (LSTM-WC) to a traditional LSTM and to an LSTM with peephole connections (LSTM-PH).

\subsection{Adding Problem and Copying Tasks}
In the adding problem~\citep{hochreiter1997long}, the input to the network consists of a series of $T$ pairs $(n_t, f_t)$, with $0\le t<T$. The first element $n_t$ is a real-valued number between 0 and 1, and $f_t$ is a corresponding marker. In the entire sequence, only two markers $f_i$ and $f_j$ are set to 1, while the others are set to 0. The goal is to predict the sum of the corresponding real-valued items $n_i + n_j$, for which $f = 1$. In our experiments, we test with $T=200$ and $T=400$, and we measure the performance using mean squared error.
%
For this experiment, the networks have hidden size $N=128$ and train for $200$ epochs. We optimize the parameters using SGD with Nesterov update rule. The learning rate is $10^{-2}$ (momentum factor $0.9$) and the batch size is $128$. We also clip the gradient norm to $1.0$.
Results are reported in Fig.~\ref{fig:plots} (top), where we plot the MSE on the test set for every epoch of training.
LSTM-WM achieves the best convergence time for $T=200$, while the final performance on this setup is similar among the three models. The effectiveness of WMCs is striking in the $T=400$ setup. In fact, the proposed model solves the adding problem around epoch $145$, while the other two architectures cannot learn the task and are stuck on the trivial solution.

In the copying task~\citep{hochreiter1997long}, the network observes a sequence of $10$ input symbols, waits for $T$ time steps (we use $T=100$ and $T=200$), and then must reproduce the same sequence as output. For this experiment, we adopt the same setup described in~\citep{arjovsky2016unitary}.
We keep the same implementation details described for the adding problem, except that we train for $500$ epochs.
In Fig.~\ref{fig:plots} (center), we plot the test accuracy achieved by the three models at each epoch. In both setups, WMCs play an important role in terms of final performance and convergence time. As in the adding problem, the performance gain given by the proposed architecture is more evident when working on longer sequences: for $T=200$, WMCs outperform peephole LSTM and vanilla LSTM by around $+25\%$ and $+40\%$.


\subsection{Permuted Sequential MNIST}
\label{sec:smnist}
The sequential MNIST (sMNIST)~\citep{le2015simple} is the sequential version of the MNIST digit recognition task~\citep{lecun1998gradient}. In this task, the image pixels are fed sequentially to the network (from left to right, and top to bottom). The permuted sequential MNIST (pMNIST) is a sequential version of the MNIST digit recognition problem in which the pixels are permuted in a random but fixed order. In both tasks, the goal is to predict the correct digit label after the last input pixel. Following the setup proposed in~\citep{arpit2018h}, we use $50$k images for training, $10$k for validation, and $10$k to test our models.
The experimental setup is as follows. We set the hidden size to $N=128$ for all the networks, and train for $200$ epochs using SGD with learning rate $10^{-2}$ and batch size $128$ (momentum $0.9$ and Nesterov update rule). We clip the gradient norms to $1.0$.

Fig.~\ref{fig:plots} (bottom) reports the mean test accuracy of the three LSTM variants for both setups. We report the standard error of the mean as a shaded area. For the sMNIST task, peephole LSTM performs slightly better than vanilla LSTM. LSTM with Working Memory Connections, instead, outperforms the competing architectures in terms of final accuracy and convergence speed. In particular, our architecture employs only 50 epochs to get above $92\%$ accuracy, while other models are still generally stuck around $65\%$ (vanilla LSTM) and $82\%$ (LSTM-PH). In this experiment, we also find out that WMCs help stabilize training. In fact, the area given by the standard error of the mean is much thicker for our approach than for the other two variants, in particular during the early stages of training.
On the pMNIST task, all the models achieve good final results, with LSTM with Working Memory Connections still being the best option.

\begin{table}[t!]
\centering
\footnotesize
\caption{Test accuracy on the sequential MNIST task.}
\setlength{\tabcolsep}{.5em}
  \begin{tabular}{lccc}
  \toprule
	 \textbf{Model} & & \textbf{sMNIST} & \textbf{pMNIST} \\
	 \midrule
	 iRNN~\citep{le2015simple} & & 97.00 & 82.00 \\
	 uRNN~\citep{arjovsky2016unitary} & & 95.10 & 91.40 \\
 	 \textit{h}-detach~\citep{arpit2018h} & & 98.50 & 92.30 \\
 	 \midrule
	 LSTM ($h=128$) & & 98.16 & 92.94 \\
	 LSTM ($h=256$) & & 97.68 & \textbf{93.97} \\
	 LSTM-PH ($h=128$) & & 98.58 & 93.25 \\
	 LSTM-PH ($h=256$) & & 98.33 & 93.40 \\
	 \textbf{LSTM-WM} ($h=128$) & & \textbf{98.63} & \textbf{93.97} \\
  \bottomrule
  \end{tabular}
  \label{table:smnist}
\end{table}

Numerical results, reported in Table~\ref{table:smnist}, confirm that our model outperforms the classic LSTM by a discrete margin ($+0.47\%$ and $+1.03\%$ on the sequential and permuted MNIST respectively). Since WMCs introduce additional learnable parameters in the LSTM structure, we also compare with vanilla and peephole LSTM with increased hidden size (256 instead of 128). Note that, in this setting, LSTM and LSTM-PH have more than $2\times$ the number of learnable parameters of LSTM-WM. Despite this, LSTM-WC achieves the best results on both tasks. It is worth noting that, while additional parameters in vanilla LSTM improves the results on pMNIST, they are not helpful in the sMNIST task. The flexibility given by WMCs, instead, allows the proposed model to achieve the best result in both setups.
Always in Table~\ref{table:smnist}, we compare with two state-of-the-art RNNs~\citep{le2015simple,arjovsky2016unitary}, and with a training algorithm for LSTM~\citep{arpit2018h}. The proposed LSTM-WC outperforms the competitors in terms of test accuracy.

\begin{table*}[t]
    \centering
    \caption{Mean test bit per character on the PTB test set. Error range indicates the standard error of the mean.}
    \footnotesize
    \begin{tabular}{lcccccc}
    \toprule
            & & \multicolumn{5}{c}{\textbf{Test Bit per Character (BPC)}} \\
    \cmidrule{3-7}
            & & \multicolumn{2}{c}{Fixed \# Params ($\sim 2.2M$)} & & \multicolumn{2}{c}{Fixed \# Hidden Units ($512$)} \\
    \textbf{Model}   & & $T_{PTB}=150$ & $T_{PTB}=300$ & & $T_{PTB}=150$ & $T_{PTB}=300$ \\
    \cmidrule{1-1}
    \cmidrule{3-4}
    \cmidrule{6-7}
    LSTM    & & 1.334 $\pm$ 0.0006 & 1.343 $\pm$ 0.0004 & & 1.386 $\pm$ 0.0005 & 1.395 $\pm$ 0.0005 \\
    LSTM-PH & & 1.339 $\pm$ 0.0048 & 1.343 $\pm$ 0.0009 & & 1.383 $\pm$ 0.0004 & 1.394 $\pm$ 0.0005 \\
    \textbf{LSTM-WM} & & \textbf{1.299 $\pm$ 0.0005} & \textbf{1.302 $\pm$ 0.0008} & & \textbf{1.299 $\pm$ 0.0005} & \textbf{1.302 $\pm$ 0.0008} \\
    \bottomrule
    \end{tabular}
    \label{tab:ptb_lm}
\end{table*}

\subsection{Penn Treebank (PTB) Character-Level Language Modeling}
Character-level language modeling requires to predict a single character at each time step given an observed sequence of text. In our experiments on the Penn Treebank (PTB) dataset~\citep{marcus1993building}, we evaluate the performance of the three different LSTM variants in terms of test mean bits per character (BPC), where lower BPC denotes better performance. We report the results in Table~\ref{tab:ptb_lm}, where we compare truncated back-propagation through time ($T_{PTB}$) over 150 and 300 steps. Since our connection introduces new learnable weights, we consider an additional setup in which we keep a fixed number of parameters for the three networks.
For this experiment, we follow the setup proposed by~\citet{merityAnalysis}, with the only exception that we employ a single LSTM layer instead of three.
%
The advantage of using Working Memory Connections is more evident for equal number of hidden units, where the proposed architecture overcomes the vanilla LSTM and peephole LSTM by a significant margin. Even when the number of parameters is fixed for all the models, LSTM-WC outperforms the competitors by $0.035$ and $0.041$ BPC for $T_{PTB} = 150$ and $T_{PTB}=300$ respectively. It is worth noting that peephole LSTM performs similarly to or even worse than vanilla LSTM on this task.

\subsection{Image Captioning}
We evaluate the performance of our LSTM with Working Memory Connections on the image captioning task, which consists of generating textual descriptions for images. We apply our approach to two different captioning models: Show and Tell~\citep{vinyals2015show} and Up-Down~\citep{anderson2018bottom}. The first model includes a single LSTM layer and does not employ attention, while the second is composed of two LSTM layers and integrates attention mechanisms over image regions. We use the Microsoft COCO dataset~\citep{lin2014microsoft} following the splits defined in~\citep{karpathy2015deep}. To represent images, we employ a global feature vector extracted from the average pooling layer of ResNet-152~\citep{he2016deep} for the Show and Tell model, and multiple feature vectors extracted from Faster R-CNN~\citep{ren2015faster} for the Up-Down architecture. We train both models with Adam optimizer~\citep{kingma2015adam} using a learning rate equal to $10^{-4}$. All other hyper-parameters are left the same as those suggested in the original papers.

\begin{table}[t]
\centering
\caption{Image captioning results on COCO test set.}
\footnotesize
\label{tab:captioning}
\setlength{\tabcolsep}{.3em}
\begin{tabular}{lccccccc}
\toprule
\textbf{Model} & & \textbf{BLEU-1} & \textbf{BLEU-4}  & \textbf{METEOR}  & \textbf{ROUGE} & \textbf{CIDEr} & \textbf{SPICE}  \\
\midrule
\textit{No Attention, ResNet-152} \\
\hspace{4mm}LSTM & & 70.9 & 27.9 & 24.4 & 51.7 & 92.0 & 17.6 \\
\hspace{4mm}GRU & & 69.5 & 26.2 & 22.7 & 50.4 & 82.3 & 15.6 \\
\hspace{4mm}LSTM-PH & & \textbf{71.4} & 27.8 & 24.3 & 51.7 & 91.1 & 17.5 \\
\hspace{4mm}\textbf{LSTM-WM} & & \textbf{71.4} & \textbf{28.3} & \textbf{24.6} & \textbf{52.4} & \textbf{94.0} & \textbf{17.8} \\
\midrule
\textit{Attention, Faster R-CNN}\\
\hspace{4mm}LSTM & & 75.9 & \textbf{36.1} & 27.4 & 56.3 & 111.9 & 20.3 \\
\hspace{4mm}GRU & & 76.0 & \textbf{36.1} & 27.0 & 56.5 & 111.0 & 20.2 \\
\hspace{4mm}LSTM-PH & & 75.8 & 35.9 & 27.3 & 56.3 & 111.5 & 20.2 \\
\hspace{4mm}\textbf{LSTM-WM} & & \textbf{76.2} & \textbf{36.1} & \textbf{27.5} & \textbf{56.5} & \textbf{112.7} & \textbf{20.4} \\
\bottomrule
\end{tabular}
\end{table}

\begin{figure}[!t]
\centering
\setlength{\tabcolsep}{.4em}
\begin{tabular}{cc}
\multicolumn{2}{c}{\scriptsize \textbf{No Attention, ResNet-152}} \\
    \includegraphics[height=0.35\linewidth]{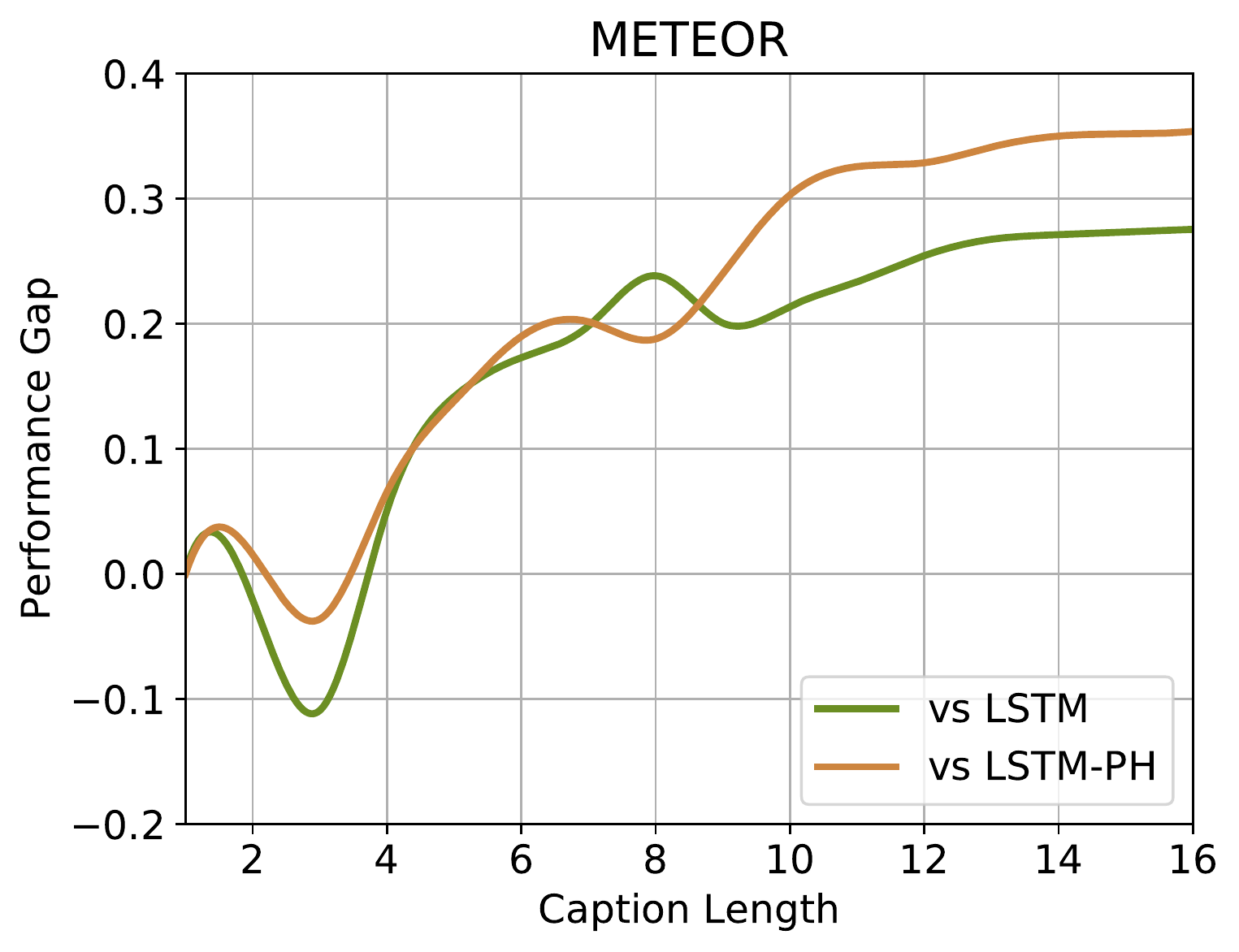} 
    &
    \includegraphics[height=0.35\linewidth]{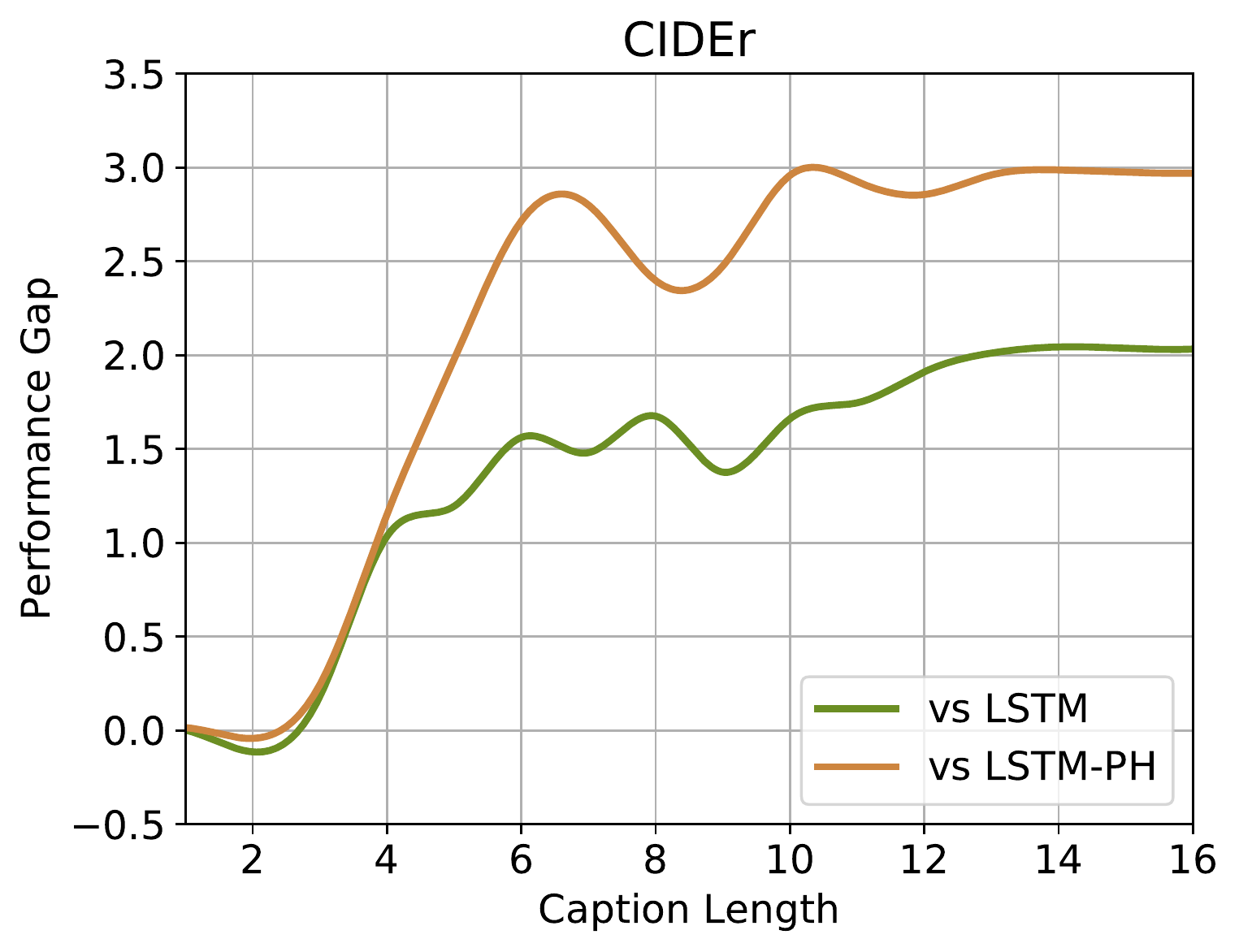} 
    \\
\multicolumn{2}{c}{\scriptsize\textbf{Attention, Faster R-CNN}} \\
    \includegraphics[height=0.35\linewidth]{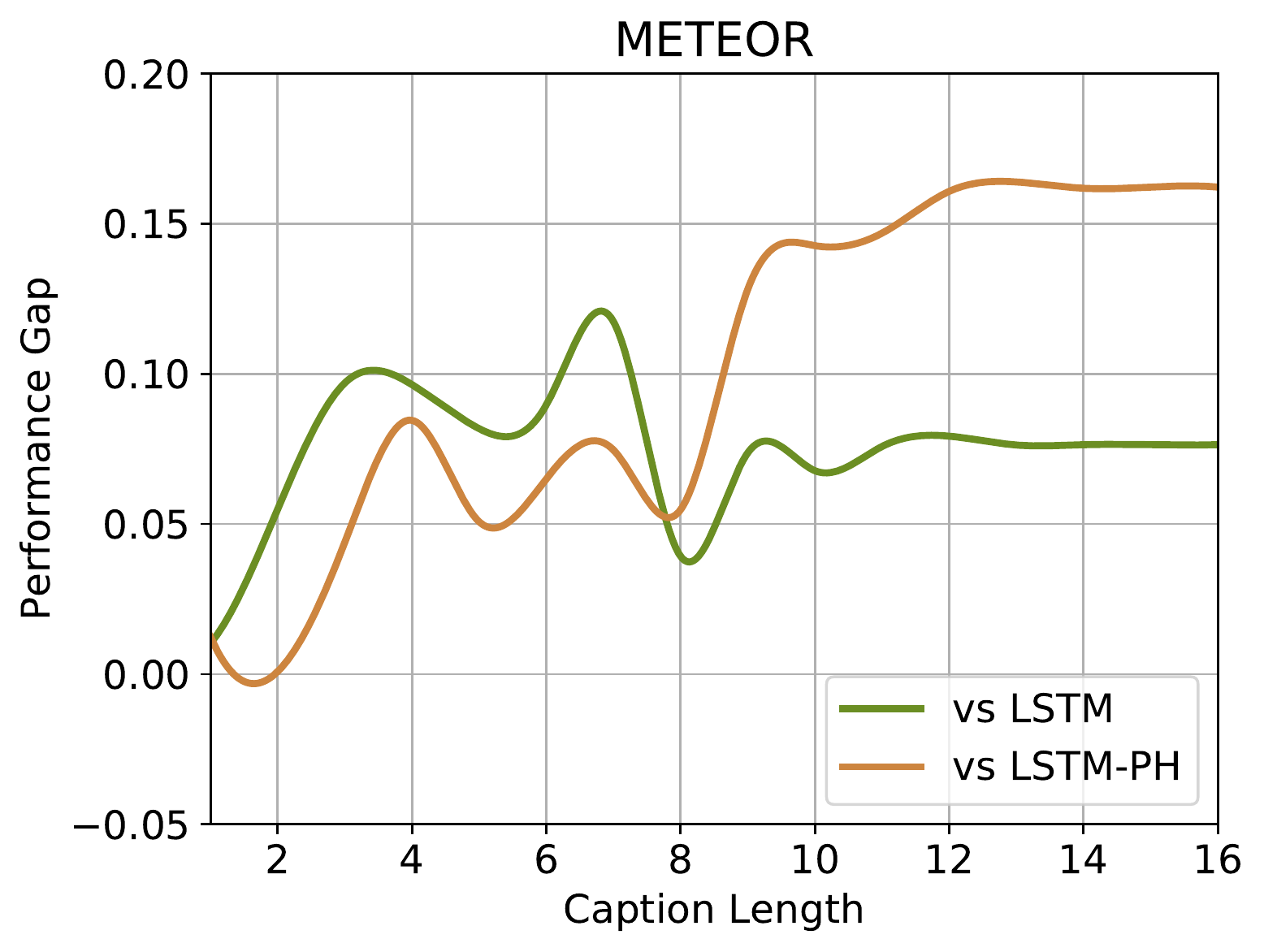}
     &  
    \includegraphics[height=0.35\linewidth]{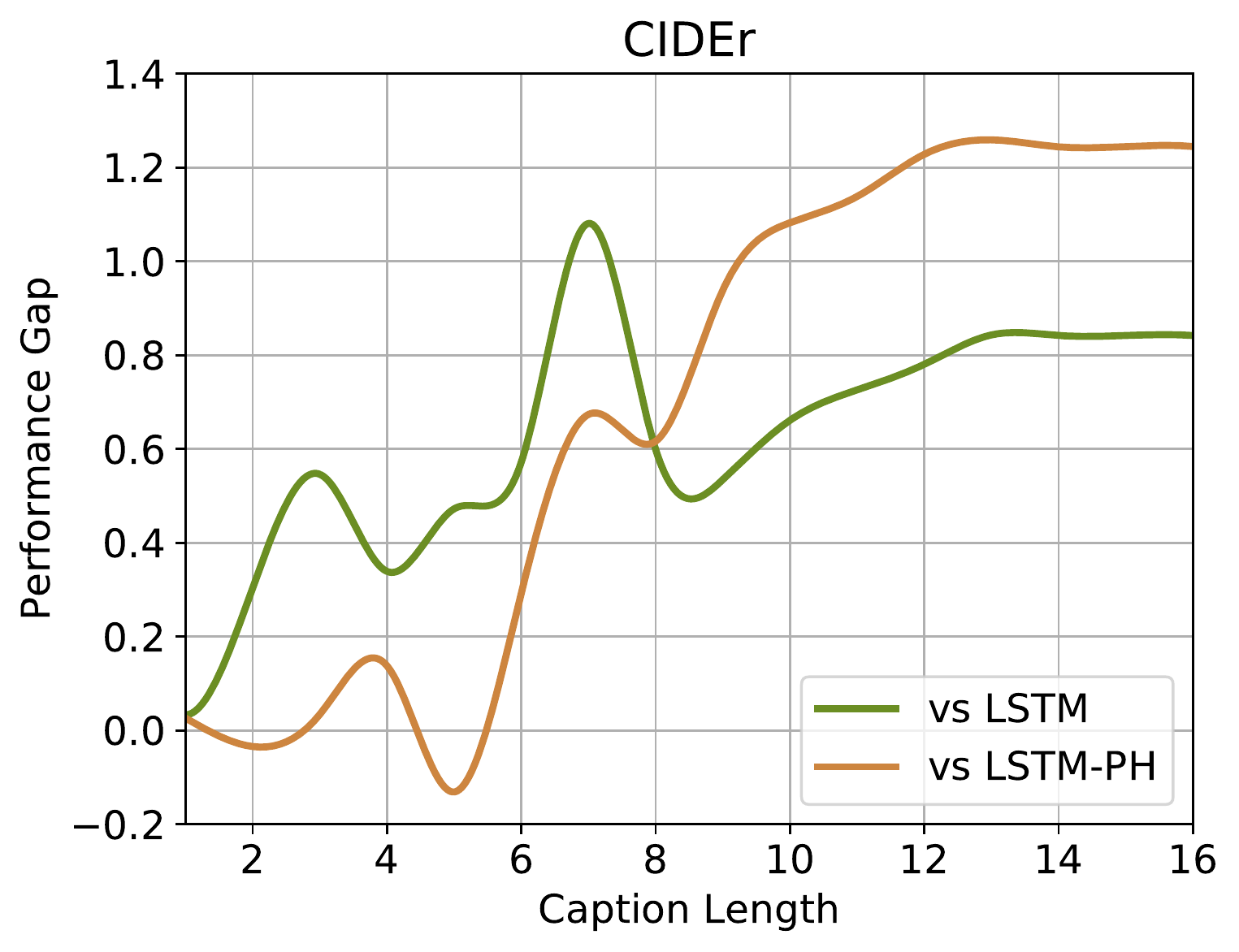}
\end{tabular}
\caption{Metric gaps on the image captioning task for increasing instruction lengths.}
\label{fig:captioning}
\end{figure}

Numerical results are reported in Table~\ref{tab:captioning} using standard captioning evaluation metrics (\ie~BLEU-1, BLEU-4~\citep{papineni2002bleu}, METEOR~\citep{banerjee2005meteor}, ROUGE~\citep{lin2004rouge}, CIDEr~\citep{vedantam2015cider}, and SPICE~\citep{anderson2016spice}). For all these, higher metric results indicate better performance, with CIDEr being the metric that best correlates with human judgment.
In both settings, our LSTM-WM outperforms traditional LSTM and LSTM-PH by a clear margin. Specifically, LSTM-WM improves the vanilla LSTM results by $2.0$ CIDEr points on the model without attention and $0.8$ CIDEr points on the model with attention over image regions, demonstrating the contribution of WMCs also for this task.
As an additional comparison, we replace the LSTM layers with GRU layers. Numerical results suggest that there is not a clear advantage in using GRUs instead of LSTMs for this task. 
In Fig.~\ref{fig:captioning}, we plot the metric gap between LSTM-WM and the two competitors in terms of METEOR and CIDEr. On the X-axis we report the length of the generated captions, meaning that we consider the first $x$ words of each predicted sentence. On the Y-axis, a $0$ value means that our proposal performs equally, \ie~has no performance gap \textit{w.r.t.} the competitor, while a higher value indicates better performance for our model. With this analysis, we aim to check whether the improvement given by WMCs can be restricted to a particular subset of the dataset. As one can observe, the metric gap generally increases with the caption length, especially \textit{w.r.t.} peephole LSTM. We can deduce that the contribution of WMCs escalates with the number of time steps.

\section{Discussion}
\label{sec:discussion}
\rev{With Working Memory Connections, we show that information stored in the LSTM cell should be accessible in the gate structure. We compare the performance of WMCs to a similar approach named peephole connections~\citep{gers2000recurrent}, and to vanilla LSTM. We find out that the structure of WMCs allows for two distinct improvements:
\begin{enumerate}
    \item \textit{A more precise control of the gates}. The multiplicative gates in the LSTM block must regulate the information flowing through the cell, but they cannot access the state of that same cell in the traditional LSTM formulation. The presence of the cell state in the multiplicative gates motivates the improvements of LSTM-WM \textit{w.r.t.}~vanilla LSTM. 
    \item \textit{Increased stability during training} compared to peephole connections. Exposing different projections of the cell state without squashing its content seems to be a critical point for the LSTM-PH. This element of novelty in our design explains why WMCs provide a boost in performance even when peepholes fail.
\end{enumerate}
As a consequence of these two improvements, WMCs incorporate the theoretical benefits of peephole connections, originally described by~\citet{gers2000recurrent}, with the training stability and versatility of vanilla LSTM.}

\rev{It is worth noting that, for tasks that do not require to access the content of the memory cell, Working Memory Connections would not probably bring any benefit in the LSTM formulation, while peepholes might still hinder the whole learning process because of unstable updates.}

\rev{At the same time, when training stacked LSTMs, the benefits given by WMCs may become less significant. We suppose that this is due to the increased complexity in the network structure, where multiple LSTM blocks can interact through the various layers.
Similarly, many architectures employ LSTMs as building blocks together with different components, and the influence of WMCs in these compound deep networks cannot be easily determined. Experiments on image captioning, proposed in this paper, partially answer this question and prove that WMCs afford a small yet existing improvement even in this scenario. However, there are many other complex tasks involving vision, language, and other modalities, that are worth investigating.}

\section{Conclusion}
\label{sec:conclusion}
A current limitation of Long Short-Term Memory Networks consists in not letting the cell state influence the gate dynamics directly. In this paper, we propose Working Memory Connections (WMCs) for LSTM, which provide an efficient way of using intra-cell knowledge inside the network. The proposed design performs noticeably better than the vanilla LSTM and overcomes important issues in previous formulations. We formally motivate this improvement as a consequence of more stable training dynamics. Experimental results reflect the theoretical benefits of the proposed approach and motivate further study in this direction. One future direction might consist in testing the efficacy of Working Memory Connections for an even wider set of tasks.

\section*{Acknowledgments}
This work has been supported by ``Fondazione di Modena'' under the project ``AI for Digital Humanities'' and by the national project ``IDEHA: Innovation for Data Elaboration in Heritage Areas'' (PON ARS01\_00421), cofunded by the Italian Ministry of University and Research.

\bibliographystyle{model2-names}
\bibliography{refs}

\begin{thebibliography}{45}
\expandafter\ifx\csname natexlab\endcsname\relax\def\natexlab#1{#1}\fi
\providecommand{\url}[1]{\texttt{#1}}
\providecommand{\href}[2]{#2}
\providecommand{\path}[1]{#1}
\providecommand{\DOIprefix}{doi:}
\providecommand{\ArXivprefix}{arXiv:}
\providecommand{\URLprefix}{URL: }
\providecommand{\Pubmedprefix}{pmid:}
\providecommand{\doi}[1]{\href{http://dx.doi.org/#1}{\path{#1}}}
\providecommand{\Pubmed}[1]{\href{pmid:#1}{\path{#1}}}
\providecommand{\bibinfo}[2]{#2}
\ifx\xfnm\relax \def\xfnm[#1]{\unskip,\space#1}\fi
\bibitem[{Anderson et~al.(2016)Anderson, Fernando, Johnson and
  Gould}]{anderson2016spice}
\bibinfo{author}{Anderson, P.}, \bibinfo{author}{Fernando, B.},
  \bibinfo{author}{Johnson, M.}, \bibinfo{author}{Gould, S.},
  \bibinfo{year}{2016}.
\newblock \bibinfo{title}{{SPICE: Semantic Propositional Image Caption
  Evaluation}}, in: \bibinfo{booktitle}{Proceedings of the European Conference
  on Computer Vision}.
\bibitem[{Anderson et~al.(2018)Anderson, He, Buehler, Teney, Johnson, Gould and
  Zhang}]{anderson2018bottom}
\bibinfo{author}{Anderson, P.}, \bibinfo{author}{He, X.},
  \bibinfo{author}{Buehler, C.}, \bibinfo{author}{Teney, D.},
  \bibinfo{author}{Johnson, M.}, \bibinfo{author}{Gould, S.},
  \bibinfo{author}{Zhang, L.}, \bibinfo{year}{2018}.
\newblock \bibinfo{title}{Bottom-up and top-down attention for image captioning
  and visual question answering}, in: \bibinfo{booktitle}{Proceedings of the
  IEEE/CVF Conference on Computer Vision and Pattern Recognition}.
\bibitem[{Arjovsky et~al.(2016)Arjovsky, Shah and Bengio}]{arjovsky2016unitary}
\bibinfo{author}{Arjovsky, M.}, \bibinfo{author}{Shah, A.},
  \bibinfo{author}{Bengio, Y.}, \bibinfo{year}{2016}.
\newblock \bibinfo{title}{Unitary evolution recurrent neural networks}, in:
  \bibinfo{booktitle}{Proceedings of the International Conference on Machine
  Learning}.
\bibitem[{Arpit et~al.(2019)Arpit, Kanuparthi, Kerg, Ke, Mitliagkas and
  Bengio}]{arpit2018h}
\bibinfo{author}{Arpit, D.}, \bibinfo{author}{Kanuparthi, B.},
  \bibinfo{author}{Kerg, G.}, \bibinfo{author}{Ke, N.R.},
  \bibinfo{author}{Mitliagkas, I.}, \bibinfo{author}{Bengio, Y.},
  \bibinfo{year}{2019}.
\newblock \bibinfo{title}{{h-detach: Modifying the LSTM Gradient Towards Better
  Optimization}}, in: \bibinfo{booktitle}{Proceedings of the International
  Conference on Learning Representations}.
\bibitem[{Bahdanau et~al.(2015)Bahdanau, Cho and Bengio}]{bahdanau2015neural}
\bibinfo{author}{Bahdanau, D.}, \bibinfo{author}{Cho, K.},
  \bibinfo{author}{Bengio, Y.}, \bibinfo{year}{2015}.
\newblock \bibinfo{title}{{Neural Machine Translation by Jointly Learning to
  Align and Translate}}, in: \bibinfo{booktitle}{Proceedings of the
  International Conference on Learning Representations}.
\bibitem[{Banerjee and Lavie(2005)}]{banerjee2005meteor}
\bibinfo{author}{Banerjee, S.}, \bibinfo{author}{Lavie, A.},
  \bibinfo{year}{2005}.
\newblock \bibinfo{title}{{METEOR: An automatic metric for MT evaluation with
  improved correlation with human judgments}}, in:
  \bibinfo{booktitle}{Proceedings of the Annual Meeting on Association for
  Computational Linguistics Workshops}.
\bibitem[{Baraldi et~al.(2017)Baraldi, Grana and
  Cucchiara}]{baraldi2017hierarchical}
\bibinfo{author}{Baraldi, L.}, \bibinfo{author}{Grana, C.},
  \bibinfo{author}{Cucchiara, R.}, \bibinfo{year}{2017}.
\newblock \bibinfo{title}{Hierarchical boundary-aware neural encoder for video
  captioning}, in: \bibinfo{booktitle}{Proceedings of the IEEE/CVF Conference
  on Computer Vision and Pattern Recognition}.
\bibitem[{Bengio et~al.(1993)Bengio, Frasconi and Simard}]{bengio1993problem}
\bibinfo{author}{Bengio, Y.}, \bibinfo{author}{Frasconi, P.},
  \bibinfo{author}{Simard, P.}, \bibinfo{year}{1993}.
\newblock \bibinfo{title}{The problem of learning long-term dependencies in
  recurrent networks}, in: \bibinfo{booktitle}{Proceedings of the International
  Joint Conference on Neural Networks}.
\bibitem[{Bengio et~al.(1994)Bengio, Simard and Frasconi}]{bengio1994learning}
\bibinfo{author}{Bengio, Y.}, \bibinfo{author}{Simard, P.},
  \bibinfo{author}{Frasconi, P.}, \bibinfo{year}{1994}.
\newblock \bibinfo{title}{Learning long-term dependencies with gradient descent
  is difficult}.
\newblock \bibinfo{journal}{IEEE Trans. on Neural Networks}
  \bibinfo{volume}{5}, \bibinfo{pages}{157--166}.
\bibitem[{Cho et~al.(2014a)Cho, Van~Merri{\"e}nboer, Bahdanau and
  Bengio}]{cho2014properties}
\bibinfo{author}{Cho, K.}, \bibinfo{author}{Van~Merri{\"e}nboer, B.},
  \bibinfo{author}{Bahdanau, D.}, \bibinfo{author}{Bengio, Y.},
  \bibinfo{year}{2014}a.
\newblock \bibinfo{title}{{On the Properties of Neural Machine Translation:
  Encoder-Decoder Approaches}}.
\newblock \bibinfo{journal}{arXiv preprint arXiv:1409.1259} .
\bibitem[{Cho et~al.(2014b)Cho, Van~Merri{\"e}nboer, Gulcehre, Bahdanau,
  Bougares, Schwenk and Bengio}]{cho2014LearningPR}
\bibinfo{author}{Cho, K.}, \bibinfo{author}{Van~Merri{\"e}nboer, B.},
  \bibinfo{author}{Gulcehre, C.}, \bibinfo{author}{Bahdanau, D.},
  \bibinfo{author}{Bougares, F.}, \bibinfo{author}{Schwenk, H.},
  \bibinfo{author}{Bengio, Y.}, \bibinfo{year}{2014}b.
\newblock \bibinfo{title}{{Learning Phrase Representations using RNN
  Encoder-Decoder for Statistical Machine Translation}}, in:
  \bibinfo{booktitle}{Proceedings of the Conference on Empirical Methods in
  Natural Language Processing}.
\bibitem[{Chung et~al.(2014)Chung, Gulcehre, Cho and
  Bengio}]{chung2014empirical}
\bibinfo{author}{Chung, J.}, \bibinfo{author}{Gulcehre, C.},
  \bibinfo{author}{Cho, K.}, \bibinfo{author}{Bengio, Y.},
  \bibinfo{year}{2014}.
\newblock \bibinfo{title}{{Empirical Evaluation of Gated Recurrent Neural
  Networks on Sequence Modeling}}.
\newblock \bibinfo{journal}{arXiv preprint arXiv:1412.3555} .
\bibitem[{Devlin et~al.(2018)Devlin, Chang, Lee and Toutanova}]{devlin2018bert}
\bibinfo{author}{Devlin, J.}, \bibinfo{author}{Chang, M.W.},
  \bibinfo{author}{Lee, K.}, \bibinfo{author}{Toutanova, K.},
  \bibinfo{year}{2018}.
\newblock \bibinfo{title}{{BERT: Pre-training of deep bidirectional
  transformers for language understanding}}, in:
  \bibinfo{booktitle}{Proceedings of the Annual Conference of the North
  American Chapter of the Association for Computational Linguistics}.
\bibitem[{Elman(1990)}]{Elman1990FindingSI}
\bibinfo{author}{Elman, J.L.}, \bibinfo{year}{1990}.
\newblock \bibinfo{title}{{Finding Structure in Time}}.
\newblock \bibinfo{journal}{Cognitive Science} \bibinfo{volume}{14},
  \bibinfo{pages}{179--211}.
\bibitem[{Ericsson and Kintsch(1995)}]{ericsson1995long}
\bibinfo{author}{Ericsson, K.A.}, \bibinfo{author}{Kintsch, W.},
  \bibinfo{year}{1995}.
\newblock \bibinfo{title}{{Long-Term Working Memory}}.
\newblock \bibinfo{journal}{Psychological Review} \bibinfo{volume}{102},
  \bibinfo{pages}{211}.
\bibitem[{Gers and Schmidhuber(2000)}]{gers2000recurrent}
\bibinfo{author}{Gers, F.A.}, \bibinfo{author}{Schmidhuber, J.},
  \bibinfo{year}{2000}.
\newblock \bibinfo{title}{Recurrent nets that time and count}, in:
  \bibinfo{booktitle}{Proceedings of the International Joint Conference on
  Neural Networks}.
\bibitem[{Gers et~al.(2000)Gers, Schmidhuber and Cummins}]{gers2000forget}
\bibinfo{author}{Gers, F.A.}, \bibinfo{author}{Schmidhuber, J.},
  \bibinfo{author}{Cummins, F.}, \bibinfo{year}{2000}.
\newblock \bibinfo{title}{{Learning to Forget: Continual Prediction with
  LSTM}}.
\newblock \bibinfo{journal}{Neural Computation} \bibinfo{volume}{12},
  \bibinfo{pages}{2451--2471}.
\bibitem[{Graves(2013)}]{graves2013generating}
\bibinfo{author}{Graves, A.}, \bibinfo{year}{2013}.
\newblock \bibinfo{title}{Generating sequences with recurrent neural networks}.
\newblock \bibinfo{journal}{arXiv preprint arXiv:1308.0850} .
\bibitem[{Graves et~al.(2013)Graves, Mohamed and Hinton}]{graves2013speech}
\bibinfo{author}{Graves, A.}, \bibinfo{author}{Mohamed, A.R.},
  \bibinfo{author}{Hinton, G.}, \bibinfo{year}{2013}.
\newblock \bibinfo{title}{Speech recognition with deep recurrent neural
  networks}, in: \bibinfo{booktitle}{Proceedings of the IEEE International
  Conference on Acoustics, Speech, and Signal Processing}.
\bibitem[{Greff et~al.(2017)Greff, Srivastava, Koutn{\'i}k, Steunebrink and
  Schmidhuber}]{Greff2017LSTMAS}
\bibinfo{author}{Greff, K.}, \bibinfo{author}{Srivastava, R.K.},
  \bibinfo{author}{Koutn{\'i}k, J.}, \bibinfo{author}{Steunebrink, B.},
  \bibinfo{author}{Schmidhuber, J.}, \bibinfo{year}{2017}.
\newblock \bibinfo{title}{{LSTM: A Search Space Odyssey}}.
\newblock \bibinfo{journal}{IEEE Transactions on Neural Networks and Learning
  Systems} \bibinfo{volume}{28}, \bibinfo{pages}{2222--2232}.
\bibitem[{He et~al.(2016)He, Zhang, Ren and Sun}]{he2016deep}
\bibinfo{author}{He, K.}, \bibinfo{author}{Zhang, X.}, \bibinfo{author}{Ren,
  S.}, \bibinfo{author}{Sun, J.}, \bibinfo{year}{2016}.
\newblock \bibinfo{title}{Deep residual learning for image recognition}, in:
  \bibinfo{booktitle}{Proceedings of the IEEE/CVF Conference on Computer Vision
  and Pattern Recognition}.
\bibitem[{Hernandez(2018)}]{hernandez2018neuroethics}
\bibinfo{author}{Hernandez, R.F.}, \bibinfo{year}{2018}.
\newblock \bibinfo{title}{{Neuroethics, Nootropics, Neuroenhancement: The
  Ethical Case Against Pharmacological Enhancements}}. volume
  \bibinfo{volume}{109}.
\newblock \bibinfo{publisher}{LIT Verlag M{\"u}nster}.
\bibitem[{Hochreiter(1991)}]{hochreiter1991untersuchungen}
\bibinfo{author}{Hochreiter, S.}, \bibinfo{year}{1991}.
\newblock \bibinfo{title}{{Untersuchungen zu dynamischen neuronalen Netzen}}.
\newblock \bibinfo{journal}{Diploma, Technische Universit{\"a}t M{\"u}nchen} .
\bibitem[{Hochreiter and Schmidhuber(1997)}]{hochreiter1997long}
\bibinfo{author}{Hochreiter, S.}, \bibinfo{author}{Schmidhuber, J.},
  \bibinfo{year}{1997}.
\newblock \bibinfo{title}{{Long Short-Term Memory}}.
\newblock \bibinfo{journal}{Neural Computation} \bibinfo{volume}{9},
  \bibinfo{pages}{1735--1780}.
\bibitem[{Jing et~al.(2019)Jing, Gulcehre, Peurifoy, Shen, Tegmark, Soljacic
  and Bengio}]{jing2019gated}
\bibinfo{author}{Jing, L.}, \bibinfo{author}{Gulcehre, C.},
  \bibinfo{author}{Peurifoy, J.}, \bibinfo{author}{Shen, Y.},
  \bibinfo{author}{Tegmark, M.}, \bibinfo{author}{Soljacic, M.},
  \bibinfo{author}{Bengio, Y.}, \bibinfo{year}{2019}.
\newblock \bibinfo{title}{{Gated Orthogonal Recurrent Units: On Learning to
  Forget}}.
\newblock \bibinfo{journal}{Neural Computation} \bibinfo{volume}{31},
  \bibinfo{pages}{765--783}.
\bibitem[{Karpathy and Fei-Fei(2015)}]{karpathy2015deep}
\bibinfo{author}{Karpathy, A.}, \bibinfo{author}{Fei-Fei, L.},
  \bibinfo{year}{2015}.
\newblock \bibinfo{title}{Deep visual-semantic alignments for generating image
  descriptions}, in: \bibinfo{booktitle}{Proceedings of the IEEE/CVF Conference
  on Computer Vision and Pattern Recognition}.
\bibitem[{Kingma and Ba(2015)}]{kingma2015adam}
\bibinfo{author}{Kingma, D.}, \bibinfo{author}{Ba, J.}, \bibinfo{year}{2015}.
\newblock \bibinfo{title}{Adam: a method for stochastic optimization}, in:
  \bibinfo{booktitle}{Proceedings of the International Conference on Learning
  Representations}.
\bibitem[{Le et~al.(2015)Le, Jaitly and Hinton}]{le2015simple}
\bibinfo{author}{Le, Q.V.}, \bibinfo{author}{Jaitly, N.},
  \bibinfo{author}{Hinton, G.E.}, \bibinfo{year}{2015}.
\newblock \bibinfo{title}{{A Simple Way to Initialize Recurrent Networks of
  Rectified Linear Units}}.
\newblock \bibinfo{journal}{arXiv preprint arXiv:1504.00941} .
\bibitem[{LeCun et~al.(1998)LeCun, Bottou, Bengio and
  Haffner}]{lecun1998gradient}
\bibinfo{author}{LeCun, Y.}, \bibinfo{author}{Bottou, L.},
  \bibinfo{author}{Bengio, Y.}, \bibinfo{author}{Haffner, P.},
  \bibinfo{year}{1998}.
\newblock \bibinfo{title}{Gradient-based learning applied to document
  recognition}.
\newblock \bibinfo{journal}{Proceedings of the IEEE} \bibinfo{volume}{86},
  \bibinfo{pages}{2278--2324}.
\bibitem[{Li et~al.(2018)Li, Gavrilyuk, Gavves, Jain and Snoek}]{LI201841}
\bibinfo{author}{Li, Z.}, \bibinfo{author}{Gavrilyuk, K.},
  \bibinfo{author}{Gavves, E.}, \bibinfo{author}{Jain, M.},
  \bibinfo{author}{Snoek, C.G.}, \bibinfo{year}{2018}.
\newblock \bibinfo{title}{Videolstm convolves, attends and flows for action
  recognition}.
\newblock \bibinfo{journal}{Computer Vision and Image Understanding}
  \bibinfo{volume}{166}, \bibinfo{pages}{41 -- 50}.
\bibitem[{Lin(2004)}]{lin2004rouge}
\bibinfo{author}{Lin, C.Y.}, \bibinfo{year}{2004}.
\newblock \bibinfo{title}{Rouge: A package for automatic evaluation of
  summaries}, in: \bibinfo{booktitle}{Proceedings of the Annual Meeting on
  Association for Computational Linguistics Workshops}.
\bibitem[{Lin et~al.(2014)Lin, Maire, Belongie, Hays, Perona, Ramanan,
  Doll{\'a}r and Zitnick}]{lin2014microsoft}
\bibinfo{author}{Lin, T.Y.}, \bibinfo{author}{Maire, M.},
  \bibinfo{author}{Belongie, S.}, \bibinfo{author}{Hays, J.},
  \bibinfo{author}{Perona, P.}, \bibinfo{author}{Ramanan, D.},
  \bibinfo{author}{Doll{\'a}r, P.}, \bibinfo{author}{Zitnick, C.L.},
  \bibinfo{year}{2014}.
\newblock \bibinfo{title}{{Microsoft COCO: Common Objects in Context}}, in:
  \bibinfo{booktitle}{Proceedings of the European Conference on Computer
  Vision}.
\bibitem[{Liu et~al.(2020)Liu, Hao, Zhang and Zhang}]{liu2020simplified}
\bibinfo{author}{Liu, Y.}, \bibinfo{author}{Hao, X.}, \bibinfo{author}{Zhang,
  B.}, \bibinfo{author}{Zhang, Y.}, \bibinfo{year}{2020}.
\newblock \bibinfo{title}{Simplified long short-term memory model for robust
  and fast prediction}.
\newblock \bibinfo{journal}{Pattern Recognition Letters} \bibinfo{volume}{136},
  \bibinfo{pages}{81--86}.
\bibitem[{Marcus and Marcinkiewicz(1993)}]{marcus1993building}
\bibinfo{author}{Marcus, M.P.}, \bibinfo{author}{Marcinkiewicz, M.A.},
  \bibinfo{year}{1993}.
\newblock \bibinfo{title}{{Building a Large Annotated Corpus of English: The
  Penn Treebank}}.
\newblock \bibinfo{journal}{Computational Linguistics} \bibinfo{volume}{19},
  \bibinfo{pages}{313--330}.
\bibitem[{Merity et~al.(2018)Merity, Keskar and Socher}]{merityAnalysis}
\bibinfo{author}{Merity, S.}, \bibinfo{author}{Keskar, N.S.},
  \bibinfo{author}{Socher, R.}, \bibinfo{year}{2018}.
\newblock \bibinfo{title}{{An Analysis of Neural Language Modeling at Multiple
  Scales}}.
\newblock \bibinfo{journal}{arXiv preprint arXiv:1803.08240} .
\bibitem[{Papineni et~al.(2002)Papineni, Roukos, Ward and
  Zhu}]{papineni2002bleu}
\bibinfo{author}{Papineni, K.}, \bibinfo{author}{Roukos, S.},
  \bibinfo{author}{Ward, T.}, \bibinfo{author}{Zhu, W.J.},
  \bibinfo{year}{2002}.
\newblock \bibinfo{title}{{BLEU: A Method for Automatic Evaluation of Machine
  Translation}}, in: \bibinfo{booktitle}{Proceedings of the Annual Meeting on
  Association for Computational Linguistics}.
\bibitem[{Pascanu et~al.(2013)Pascanu, Mikolov and
  Bengio}]{pascanu2013difficulty}
\bibinfo{author}{Pascanu, R.}, \bibinfo{author}{Mikolov, T.},
  \bibinfo{author}{Bengio, Y.}, \bibinfo{year}{2013}.
\newblock \bibinfo{title}{On the difficulty of training recurrent neural
  networks}, in: \bibinfo{booktitle}{Proceedings of the International
  Conference on Machine Learning}.
\bibitem[{Ren et~al.(2015)Ren, He, Girshick and Sun}]{ren2015faster}
\bibinfo{author}{Ren, S.}, \bibinfo{author}{He, K.}, \bibinfo{author}{Girshick,
  R.}, \bibinfo{author}{Sun, J.}, \bibinfo{year}{2015}.
\newblock \bibinfo{title}{{Faster R-CNN: Towards real-time object detection
  with region proposal networks}}, in: \bibinfo{booktitle}{Advances in Neural
  Information Processing Systems}.
\bibitem[{Rumelhart et~al.(1986)Rumelhart, Hinton and
  Williams}]{Rumelhart1986LearningRB}
\bibinfo{author}{Rumelhart, D.E.}, \bibinfo{author}{Hinton, G.E.},
  \bibinfo{author}{Williams, R.J.}, \bibinfo{year}{1986}.
\newblock \bibinfo{title}{Learning representations by back-propagating errors}.
\newblock \bibinfo{journal}{Nature} \bibinfo{volume}{323},
  \bibinfo{pages}{533--536}.
\bibitem[{Sutskever et~al.(2014)Sutskever, Vinyals and
  Le}]{sutskever2014sequence}
\bibinfo{author}{Sutskever, I.}, \bibinfo{author}{Vinyals, O.},
  \bibinfo{author}{Le, Q.V.}, \bibinfo{year}{2014}.
\newblock \bibinfo{title}{Sequence to sequence learning with neural networks},
  in: \bibinfo{booktitle}{Advances in Neural Information Processing Systems}.
\bibitem[{Vaswani et~al.(2017)Vaswani, Shazeer, Parmar, Uszkoreit, Jones,
  Gomez, Kaiser and Polosukhin}]{vaswani2017attention}
\bibinfo{author}{Vaswani, A.}, \bibinfo{author}{Shazeer, N.},
  \bibinfo{author}{Parmar, N.}, \bibinfo{author}{Uszkoreit, J.},
  \bibinfo{author}{Jones, L.}, \bibinfo{author}{Gomez, A.N.},
  \bibinfo{author}{Kaiser, {\L}.}, \bibinfo{author}{Polosukhin, I.},
  \bibinfo{year}{2017}.
\newblock \bibinfo{title}{Attention is all you need}, in:
  \bibinfo{booktitle}{Advances in Neural Information Processing Systems}.
\bibitem[{Vedantam et~al.(2015)Vedantam, Lawrence~Zitnick and
  Parikh}]{vedantam2015cider}
\bibinfo{author}{Vedantam, R.}, \bibinfo{author}{Lawrence~Zitnick, C.},
  \bibinfo{author}{Parikh, D.}, \bibinfo{year}{2015}.
\newblock \bibinfo{title}{{CIDEr: Consensus-based Image Description
  Evaluation}}, in: \bibinfo{booktitle}{Proceedings of the IEEE/CVF Conference
  on Computer Vision and Pattern Recognition}.
\bibitem[{Vinyals et~al.(2015)Vinyals, Toshev, Bengio and
  Erhan}]{vinyals2015show}
\bibinfo{author}{Vinyals, O.}, \bibinfo{author}{Toshev, A.},
  \bibinfo{author}{Bengio, S.}, \bibinfo{author}{Erhan, D.},
  \bibinfo{year}{2015}.
\newblock \bibinfo{title}{{Show and Tell: A Neural Image Caption Generator}},
  in: \bibinfo{booktitle}{Proceedings of the IEEE/CVF Conference on Computer
  Vision and Pattern Recognition}.
\bibitem[{Xiao et~al.(2020)Xiao, Xu and Shi}]{XIAO2020173}
\bibinfo{author}{Xiao, H.}, \bibinfo{author}{Xu, J.}, \bibinfo{author}{Shi,
  J.}, \bibinfo{year}{2020}.
\newblock \bibinfo{title}{Exploring diverse and fine-grained caption for video
  by incorporating convolutional architecture into lstm-based model}.
\newblock \bibinfo{journal}{Pattern Recognition Letters} \bibinfo{volume}{129},
  \bibinfo{pages}{173 -- 180}.
\bibitem[{Xu et~al.(2015)Xu, Ba, Kiros, Cho, Courville, Salakhudinov, Zemel and
  Bengio}]{xu2015show}
\bibinfo{author}{Xu, K.}, \bibinfo{author}{Ba, J.}, \bibinfo{author}{Kiros,
  R.}, \bibinfo{author}{Cho, K.}, \bibinfo{author}{Courville, A.},
  \bibinfo{author}{Salakhudinov, R.}, \bibinfo{author}{Zemel, R.},
  \bibinfo{author}{Bengio, Y.}, \bibinfo{year}{2015}.
\newblock \bibinfo{title}{{Show, Attend and Tell: Neural Image Caption
  Generation with Visual Attention}}, in: \bibinfo{booktitle}{Proceedings of
  the International Conference on Machine Learning}.

\end{thebibliography}



\end{document}